\documentclass[sigconf]{acmart}
\usepackage{ulem}
\usepackage{algorithm}
\usepackage{algorithmic}
\usepackage{subfigure}
\usepackage{enumerate}
\usepackage{multirow}
\renewcommand{\algorithmicensure}{\textbf{Output:}}

\renewcommand{\paragraph}[1]{\vspace{1ex}\noindent\textbf{#1}.}

\AtBeginDocument{%
  \providecommand\BibTeX{{%
    \normalfont B\kern-0.5em{\scshape i\kern-0.25em b}\kern-0.8em\TeX}}}

\copyrightyear{2023}
\acmYear{2023}
\setcopyright{acmlicensed}
\acmConference[MM '23] {Proceedings of the 31st ACM International Conference on Multimedia}{October 29--November 3, 2023}{Ottawa, ON, Canada.}
\acmBooktitle{Proceedings of the 31st ACM International Conference on Multimedia (MM '23), October 29--November 3, 2023, Ottawa, ON, Canada}
\acmPrice{15.00}
\acmISBN{979-8-4007-0108-5/23/10}
\acmDOI{10.1145/3581783.3611981}
\begin{document}

    \title{Scalable Incomplete Multi-View Clustering with Structure Alignment}

\author{Yi Wen}
\email{wenyi21@nudt.edu.com}
\affiliation{%
  \institution{National University of Defense
Technology}
  \city{Changsha}
  \country{China}
}

\author{Siwei Wang}
\authornote{Corresponding author}
\email{wangsiwei13@nudt.edu.com}
\affiliation{%
  \institution{Intelligent Game and Decision Lab}
  \city{Beijing}
  \country{China}
}

\author{Ke Liang}
\email{liangke200694@126.com}
\affiliation{%
  \institution{National University of Defense
Technology}
  \city{Changsha}
  \country{China}
}

\author{Weixuan Liang}
\email{weixuanliang@nudt.edu.cn}
\affiliation{%
  \institution{National University of Defense
Technology}
  \city{Changsha}
  \country{China}
}

\author{Xinhang Wan}
\email{wanxinhang@nudt.edu.cn}
\affiliation{%
  \institution{National University of Defense
Technology}
  \city{Changsha}
  \country{China}
}

\author{Xinwang Liu}
\authornotemark[1]
\email{xinwangliu@nudt.edu.cn}
\affiliation{%
  \institution{National University of Defense
Technology}
  \city{Changsha}
  \country{China}
}

\author{Suyuan Liu}
\email{suyuanliu@nudt.edu.cn}
\affiliation{%
  \institution{National University of Defense
Technology}
  \city{Changsha}
  \country{China}
}

\author{Jiyuan Liu}
\email{liujiyuan13@nudt.edu.cn}
\affiliation{%
  \institution{National University of Defense
Technology}
  \city{Changsha}
  \country{China}
}

\author{En Zhu}
\email{enzhu@nudt.edu.com}
\affiliation{%
  \institution{National University of Defense
Technology}
  \city{Changsha}
  \country{China}
}
\renewcommand{\shortauthors}{Yi Wen et al.}

\begin{abstract}
The success of existing multi-view clustering (MVC) relies on the assumption that all views are complete. However, samples are usually partially available due to data corruption or sensor malfunction, which raises the research of incomplete multi-view clustering (IMVC). Although several anchor-based IMVC methods have been proposed to process the large-scale incomplete data, they still suffer from the following drawbacks: i) Most existing approaches neglect the inter-view discrepancy and enforce cross-view representation to be consistent, which would corrupt the representation capability of the model; ii) Due to the samples disparity between different views, the learned anchor might be misaligned, which we referred as the Anchor-Unaligned Problem for Incomplete data (AUP-ID). Such the AUP-ID would cause inaccurate graph fusion and degrades clustering performance. To tackle these issues, we propose a novel incomplete anchor graph learning framework termed Scalable Incomplete Multi-View Clustering with Structure Alignment (SIMVC-SA). Specially, we construct the view-specific anchor graph to capture the complementary information from different views. In order to solve the AUP-ID, we propose a novel structure alignment module to refine the cross-view anchor correspondence. Meanwhile, the anchor graph construction and alignment are jointly optimized in our unified framework to enhance clustering quality. Through anchor graph construction instead of full graphs, the time and space complexity of the proposed SIMVC-SA is proven to be linearly correlated with the number of samples. Extensive experiments on seven incomplete benchmark datasets demonstrate the effectiveness and efficiency of our proposed method. Our code is publicly available at \url{https://github.com/wy1019/SIMVC-SA}.
\end{abstract}




\maketitle

\section{Introduction}
Multi-view data, which consists of features extracted from objects by different sensors, have been massively generated in recent years. Multi-modal information \cite{wang2021survey, li2023dual, AKGR,li2018survey, TGC_ML,liuyue_survey,liuyue_HSAN} 
 about the data can commonly be utilized to enhance the expressive ability of the model. For instance, the same news can be described from different views, \textit{i.e.,} textual reports \cite{liang2021mka} and visual pictures \cite{nie2017auto,xu2023adaptive,zhao2022mose,10.1145/3503161.3547864}. As an essential paradigm of multi-view learning, multi-view clustering (MVC) has drawn substantial attention because of its promising capability to reveal the intrinsic data structure \cite{cai2013multi,li2019flexible}. 
In general, multi-view clustering achieves remarkable performance by learning a consensus representation by exploring consistency among diverse views \cite{kumar2011co,zhao2017multi,ren2019semi}. For instance, \citet{zhan2018multiview} optimizes the final consensus graph by imposing low-rank constraints and minimizing the discrepancies of individual graphs. 
\citet{zhang2017latent} reconstruct samples in the latent space to achieve a more precise and reliable subspace representation.

Although numerous methods have been proposed to enhance MVC in diverse ways, most of them assume that all data are fully available \cite{gao2015multi,qiang2021fast,zhan2017graph,peng2019comic,wan2023autoweighted,wan2023onestep, LiangTNNLS, LiangTKDE, ZJPACMMM}. However, samples are often partially available in real scenarios due to data corruption or sensor malfunction. For instance,  in software traffic detection, people can't use all detected software, which leads to the incompleteness of the samples in the corresponding view.
 The different sample absence between views destroys the original cross-view alignment information and enlarges the difficulty of exploring consensus and complementary information,  
  causing incomplete multi-view clustering (IMVC) a challenging problem. To tackle these issues, several IMVC methods have been proposed in previous literature. For instance,  \citet{li2014partial} learn a common potential representation from incomplete samples by non-negative matrix decomposition and $\ell_1$ regularization terms. \citet{wen2020generalized} propose a new regularization term to preserve the local geometric structure and fuse the individual incomplete graph. Although remarkable success has been made, the high time complexity hinders their application in large-scale scenarios\cite{liuyue_Dink_net}. One pioneer work, \citet{liu2022fast} efficiently reduce the algorithm complexity by utilizing the anchor graph to capture the clustering structure with incomplete views.

\begin{figure} \centering    

\includegraphics[width=0.99\linewidth]{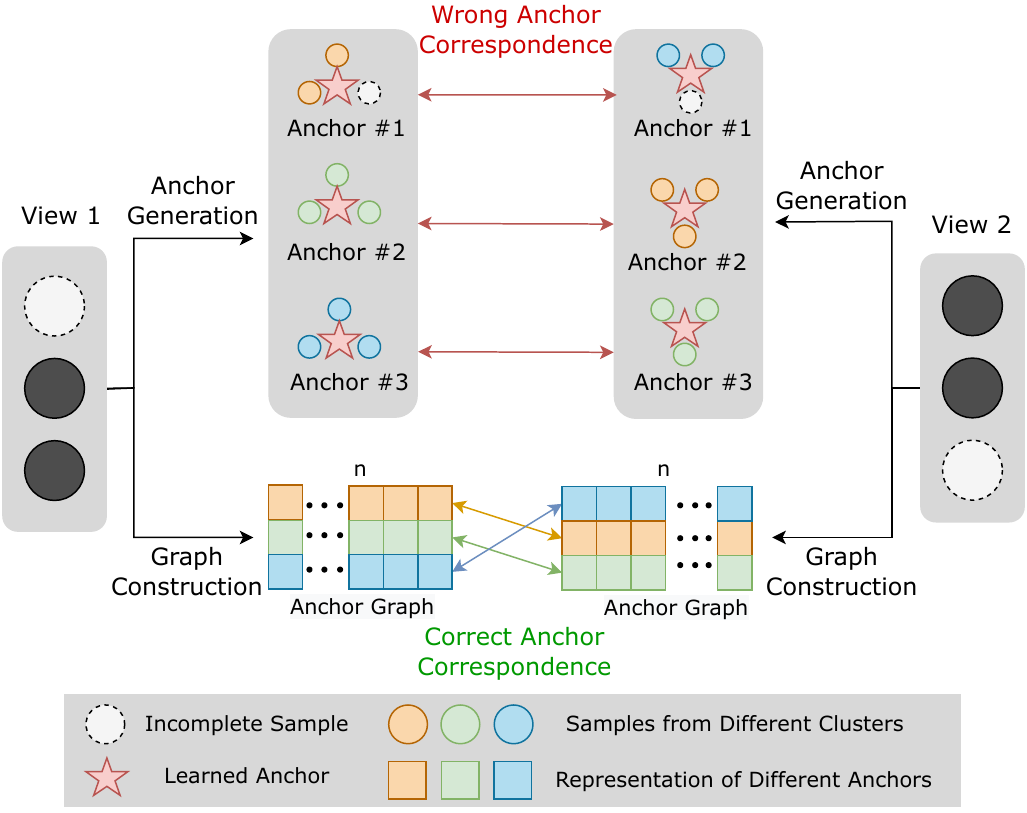}     
\caption{An example illustration of AUP-ID. Samples and representations of different colors represent samples from different clusters and representations of different anchors, respectively. With different missing samples and random anchor initialization, the anchor learned may be unaligned and leads to inaccurate correspondences.}     
\label{explanation}     
\setlength{\abovecaptionskip}{-1pt}
\end{figure}

Although widely applied in large-scale applications, the existing anchor-based IMVC methods still suffer from the following drawbacks: Firstly, most approaches neglect the inter-view discrepancy and enforce cross-view representation to be consistent, which would corrupt the representation capability of the model. Secondly, as is shown in Fig. \ref{explanation}, the sample distribution of different views might be biased due to the incomplete multi-view data, which leads to the potential misalignment between cross-view anchors, which we referred as the Anchor-Unaligned Problem for Incomplete Data (AUP-ID). Such AUP-ID would result in inaccurate graph fusion and suboptimal clustering performance. This issue for complete data has been demonstrated in \cite{wang2022align,huang2020partially} and has more significant implications for IMVC due to the incomplete cross-view alignment information. Besides, to the best of our knowledge, no generalized framework for solving AUP-ID has been proposed since generating the correct cross-view anchor correspondence under incomplete scenarios would be more difficult due to the variance of the feature dimension and available sample.
\vspace{-2pt}

To tackle these challenging issues, we propose a novel incomplete anchor graph learning framework termed Scalable Incomplete Multi-View Clustering with Structure Alignment (SIMVC-SA). Specifically, we construct the incomplete anchor graph on each view to capture the complementary information from different views. In order to address the AUP-ID, we adopt a novel structure alignment module to refine the cross-view anchor correspondence mapping adequately. Meanwhile, the anchor graph construction and alignment are jointly optimized in our unified framework to enhance clustering quality. In addition, through the anchor graph construction rather than a full pairwise graph, the time complexity of the SIMVC-SA is effectively reduced from $\mathcal{O}(n^3)$ to $\mathcal{O}(nm)$. Meanwhile, a convergent five-step alternative algorithm is designed in this paper to tackle the subsequent optimization problem. We summarize the contributions as follows:
\begin{itemize}
    \item In order to solve the Anchor-Unaligned Problem for Incomplete Data, a novel alignment module is proposed in this paper to capture the view-specific structure. With the guidance of structure information, the cross-view anchor correspondence mapping can be refined adequately.

    \item We design a novel IMVC  approach termed Scalable Incomplete Multi-View Clustering with Structure Alignment (SIM VC-SA). Different from the existing fixed anchor strategy, SIMVC-SA  learns the anchor and constructs the respective anchor graph to enhance the clustering performance.

    \item Extensive experiments on seven incomplete benchmark datasets show the effectiveness and efficiency of the proposed method. 
\end{itemize}

\vspace{-10pt}
\section{Related Work}
\subsection{Incomplete Multi-View Clustering (IMVC)}
In real scenarios, samples are often partially available due to data dropout and sensor corruption, which raises the incomplete multi-view clustering (IMVC) study \cite{yang2022robust,hu2019one,lin2022incomplete,xu2022multi,jin2023deep,wan2023fast}. The existing Incomplete Muti-View Clustering (IMVC) approaches can be roughly divided into three types:  Non-negative matrix factorization (NMF) methods \cite{zhao2016incomplete}, kernel or graph-based methods \cite{nie2016parameter,han2022incomplete}, and deep neural networks \cite{wen2021structural,zhu2019multi,wang2020icmsc}.

In general, NMF jointly decomposes the raw matrix of each view into a coefficient matrix and a basis matrix and learns a consensus matrix from the coefficient matrices with a group of adaptive view weights. 
The graph or kernel-based IMVC approach performs matrix complementation and achieves the desired clustering performance by constructing consensus graphs or kernels \cite{wang2020smoothness,liu2020multiple,ren2020consensus}.
For example, \citet{wang2019spectral} propose a novel similarity matrix padding strategy based on matrix perturbation theory. Because of the capacity to extract high-level information, deep neural networks often achieve desirable performance for solving IMVC problems \cite{wen2020cdimc,wang2018partial,lin2022dual}.   \citet{lin2021completer} design a deep IMVC model through the union of representation and cross-view data recovery.

\subsection{Graph-based IMVC Method}
Graph structures \cite{kang2020partition,khan2019approximate,wang2019gmc, Mo_AAAI_2022,liang2023structure, S2T_ML, CCGC, GCC-LDA, MGCN,liuyue_DCRN,liuyue_SCGC}, which can well describe the relationships of pairwise data, are widely adopted in the field of IMVC. 
Denoting the indicator vector $\mathbf{w}^{(v)} \in \mathbb{R}^{n_v}$ containing the index for $n_v$ available samples in the $v$-th view, we define the index matrix $\mathbf{H}_v \in \mathbb{R}^{n \times n_v}$ for $v$-th view as follows: 

 \vspace{-10pt}
$$
\mathbf{h}_{i j}^{(v)}= \begin{cases}1, & \text { if } w_j^{(v)}=i, \\ 0, & \text { otherwise. }\end{cases}
$$
\vspace{-2pt}
where $\mathbf{h}_{i j}^{(v)}$ denotes the element in $i$-th row and $j$-th column of $\mathbf{H}_v$. Then, $\mathbf{X}_v \mathbf{H}_v \in \mathbb{R}^{d_v \times n_v}$ denotes the existing data matrix of the $v$-th view.

As to the incomplete multi-view data, the subgraphs from each view may have a few blanks in the respective rows and columns because of the incomplete setting. Taking this into account, the classical graph-based IMVC paradigms \cite{li2022refining,liu2010large} could be mathematically represented in two parts:

\begin{equation}
\begin{aligned}
\label{gimvc}
\min_{\mathbf{S}_v, \mathbf{S}} &\left\|\mathbf{X}_v \mathbf{H}_v-\mathbf{X}_v \mathbf{H}_v \mathbf{S}_v\right\|_{\mathbf{F}}^2+\Psi\left(\mathbf{H}_v \mathbf{S}_v \mathbf{H}_v^{\top}, \mathbf{S}\right),\\
\text { s.t. } & \mathbf{S}_v \geq 0, \quad \mathbf{S}_v^{\top} \mathbf{1}=\mathbf{1}, \quad \mathbf{S} \geq 0,  \quad \mathbf{S}^{\top} \mathbf{1}=\mathbf{1}
\end{aligned}
\end{equation}
where $\mathbf{S}_v \in \mathbb{R}^{n_v \times n_v}$ is the view-specific subgraph, $\mathbf{S}$ indicates the similarity among all the samples. The $\Psi(\cdot)$ indicates the graph fusion process. However, $\mathcal{O}\left(v n^2\right)$ space complexity and $\mathcal{O}\left(n^3\right)$ time expenditure prevent this category of algorithms from handling large-scale incomplete multiview tasks \cite{kang2020large}. 

\subsection{Anchor-based IMVC Method}
As is shown in Eq.(\ref{gimvc}), the majority of the classical graph-based IMVC methods involve the full graph construction, which makes them suffer from $\mathcal{O}\left(n^3\right)$ time complexity. To tackle the issues, \citet{liu2022fast,li2022parameter} propose the anchor-based incomplete multi-view clustering (AIMVC). The complexity of AIMVC is effectively reduced by merely building the relationship between representative anchors and samples \cite{huang2019auto}. The classical AIMVC framework can be formulated as follows:
\begin{equation}
\begin{aligned}
\label{aimvc}
\min _{\mathbf{Z}} &\left\|\mathbf{X}_v \mathbf{H}_v-\mathbf{A}_v \mathbf{Z} \mathbf{H}_v\right\|_{\mathbf{F}}^2+\Omega\left(\mathbf{Z}\right), \\ 
& \text { s.t. } \mathbf{Z} \geq 0, \quad  \mathbf{Z}^{\top} \mathbf{1}=\mathbf{1},
\end{aligned}
\end{equation}
where $\mathbf{A}_p \in \mathbb{R}^{d_v \times m}$ denotes the anchor matrix from the $v$-th view, $\mathbf{Z}$ is the consistent anchor graph, $m$ is the number of anchors, and $\Omega$ is the regularization term. 

On the basis of the above paradigm, many methods adopt different regularization terms to enhance clustering performance \cite{liu2022fast,guo2019anchors}. However, most existing approaches overlook the inter-view discrepancy and enforce cross-view representation to be consistent, which would corrupt the representation capability of the model. Moreover,  the potential Anchor-Unaligned Problem for Incomplete Data has not been discussed in previous research. Such AUP-ID would result in inaccurate graph fusion and suboptimal clustering performance. In the next section, we will propose SIMVC-SA to tackle these issues.

\begin{figure*} \centering    
\includegraphics[width=0.9\linewidth]{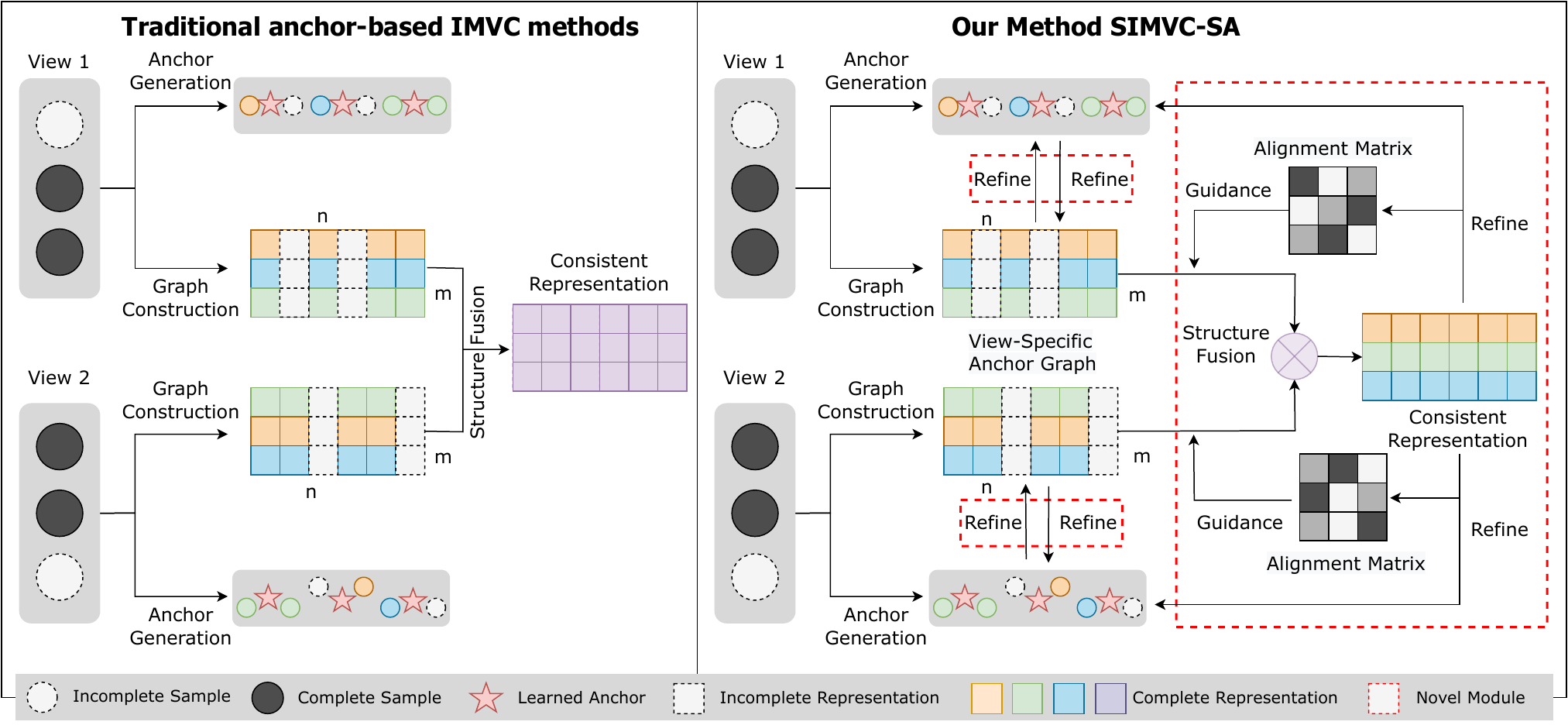}     
\caption{The framework of traditional anchor-based IMVC methods (left) and the proposed SIMVC-SA (right). Different from traditional IMVC methods, the proposed SIMVC-SA  proposes a novel structure alignment module and adopts the anchor learning strategy to efficiently enhance the clustering performance. }     
\label{framework}     
\end{figure*}

\section{Methods}
\subsection{Problem Formulation}
As mentioned before, the main challenge for solving AUP-ID is the variation of the feature dimension and available sample, which results in the anchors from different views being under different metric spaces, and we can't directly measure the distance of cross-view anchors. As a result, a question worth considering: \textbf{how to effectively refine the cross-view anchor correspondence under the incomplete scenario?} 

An intuitive method \cite{liu2022fast} to implicitly avoid anchor correspondence is to enforce cross-view anchor and the respective graph to be consistent. While such a strategy overlooks the inter-view discrepancy and corrupts the representation capability of the model. Inspired by \cite{wang2022align,huang2020partially}, we consider such principle: the correspondence probability of the anchors should be high if their corresponding structure is similar. Therefore, the original anchor correspondence problem can be transferred to the structure alignment problem, as depicted in Fig. \ref{explanation}. 
In this paper, we introduce the alignment matrix $\mathbf{P}_v$ that satisfies $\mathbf{P}_v^{\top} \mathbf{P}_v=\mathbf{I}_m$ to efficiently tackle the problem. Denoting the fusion representation as $\mathbf{F}$, and the anchor graph alignment problem can be addressed as follow:

\begin{equation} \label{aligned}
\min_{\mathbf{P}_v} \left\|\mathbf{P}_v \mathbf{Z}_v-\mathbf{F}\right\|_F^2, \quad \text {s.t.} \quad \mathbf{P}_v^{\top} \mathbf{P}_v=\mathbf{I}_m ,
\end{equation} 
where $\mathbf{Z}_v \in$ $\mathbb{R}^{m \times n}$ is the view-specific anchor graph.

Moreover, considering the traditional fixed anchor strategy relies on the quality of anchors initialization and introduces unnecessary time overhead, we adopt the anchor learning strategy to enhance our clustering performance in this paper. In summary, the proposed Scalable Incomplete Multi-View Clustering with Structure Alignment (SIMVC-SA) can be formulated as follows:

\begin{equation}
\label{my formula}
\begin{aligned}
& \min_{\substack{\boldsymbol{\gamma},\left\{\mathbf{A}_v\right\}_{v=1}^V,\\ \left\{\mathbf{Z}_v\right\}_{v=1}^V, \mathbf{P}, \mathbf{F}}}   \  \sum_{v=1}^V \gamma_v^2\left\|\mathbf{X}_v \mathbf{H}_v-\mathbf{A}_v \mathbf{Z}_v \mathbf{H}_v\right\|_F^2 \\
& \qquad \quad \ +\lambda \sum_{v=1}^V\left\|\mathbf{P}_v \mathbf{Z}_v-\mathbf{F} \right\|_F^2 + \mu \sum_{v=1}^V \left\|\mathbf{Z}_v \right\|_F^2 \\
& \text {s.t.} \boldsymbol{\gamma}^{\top} 1=1, \mathbf{A}_v^{\top} \mathbf{A}_v=\mathbf{I}_m, \mathbf{P}_v^{\top} \mathbf{P}_v=\mathbf{I}_m,  \mathbf{Z}_v \geq 0, \\
& \mathbf{Z}_v^{\top} \mathbf{1}_m=\mathbf{1}_n,  \mathbf{F} \mathbf{F}^\top = \mathbf{I}_m ,
\end{aligned}
\end{equation}
where $\mathbf{Z}_v \mathbf{H}_v$ can be considered as the similarities between $m$ anchors and $n_v$ available samples of the $v$-th view. For the sake of  making the learned anchors $\mathbf{A}_v$ and consistent representation $\mathbf{F}$ more discriminative, we impose orthogonal constraints into them that $\mathbf{A}^{\top}_v \mathbf{A}_v=\mathbf{I}_m, \mathbf{F} \mathbf{F}^\top = \mathbf{I}_m$. The learned bipartite graph $\mathbf{Z}_v$ should satisfy $\mathbf{Z}_v \geq 0$ and $\mathbf{Z}_v^{\top} \mathbf{1}=\mathbf{1}$. The $\boldsymbol{\gamma} \in \mathbb{R}^V$ captures the weight contribution of every single view to all. $\lambda$ is the trade-off to balancing the
influence between anchor graph generation and alignment term. $\mu$ is the hyperparameter of the regularization term. The framework of our SIMVC-SA is shown in Fig. \ref{framework}.

Although the Eq.\eqref{my formula} appears to be simple, we emphasize the superiority of SIMVC-SA as follows:
\begin{enumerate}[(1)]
\item \textbf{Joint Optimization Model.} Unlike the existing two-stage "aligning then clustering" strategy \cite{wang2022align}, we propose a joint alignment-clustering framework where the consistent representation $\mathbf{F}$ and the alignment matrix $\mathbf{P}_v$ can be joint optimized to enhance the final clustering performance in our model.
\item \textbf{Flexible Model with No Reference View.} Different from the FMVACC \cite{wang2022align}, which selects the first view for reference (all views align to the first view) and MvCLN \cite{yang2021partially} iteratively selects the reference view, we set a consistent representation $\mathbf{F}$ for alignment and optimize it adaptively in our mode which avoids catastrophic performance degradation when the reference view has poor quality\cite{lin2022tensor,gong2022gromov}.  
 \item  \textbf{Soft Alignment Correspondence.} The strict one-to-one mapping proposed in \cite{wang2022align} neglects the relationship between the different anchors and brings higher time expenses. Besides, it is too harsh and unreasonable to completely push away the different anchors. Recent work CLIP \cite{gao2023softclip} also noticed this problem. In the proposed method, we relax the original strict constraint to an orthogonal constraint to achieve a soft assignment while effectively reducing the time complexity of the alignment.
\end{enumerate}

\subsection{Optimization}
The optimization problem in Eq.(\ref{my formula}) is a non-convex problem when taking all variables into account. To solve this problem in the section, we develop an iterative optimization algorithm to address it. For the sake of simplifying the optimization procedure, we have that  $\mathbf{X}_v \mathbf{H}_v \mathbf{H}_v^{\top}=\mathbf{X}_v \odot \mathbf{R}_v$, where $\mathbf{R}_v = \mathbf{1}_{d_v} \mathbf{r}^{(v)} \in  \mathbb{R}^{d_v \times n}$, $
\mathbf{r}^{(v)}=[r_1^{(v)}, \cdots, r_n^{(v)}]$, where $r_i^{(v)}=\sum_{j=1}^{n_v} \mathbf{H}_{ij}^{(v)}$, $\odot$ represents the Hadamard product. With this transformation, the space
complexity drops from $\mathcal{O}(vn^2)$ to $\mathcal{O}(dn)$.

\subsubsection{Optimization of Anchor Matrices \texorpdfstring{$\left\{\mathbf{A}_v\right\}_{v=1}^V$}{}}

When $\left\{\mathbf{Z}_v\right\}_{v=1}^V$, $\left\{\mathbf{P}_v\right\}_{v=1}^V$,  $\boldsymbol{\gamma}$ and $\mathbf{F}$ are fixed, the optimization for $\left\{\mathbf{A}_v\right\}_{v=1}^V$ can be written as follows:
\begin{equation}
\begin{aligned}
\label{opt A}
&\min_{\left\{\mathbf{A}_v\right\}_{v=1}^V} \sum_{v=1}^V \gamma_v^2\left\|\mathbf{X}_v \mathbf{H}_v-\mathbf{A}_v \mathbf{Z}_v \mathbf{H}_v\right\|_F^2, \\ 
&\text { s.t. } \quad  \mathbf{A}_v^{\top} \mathbf{A}_v=\mathbf{I}_m.
\end{aligned}
\end{equation}
Considering the optimization of each $\mathbf{A}_v$ is independent of the corresponding view. Therefore, we extend the Frobenius norm with traces and remove the  irrelevant item, Eq.(\ref{opt A}) can be formulated as:

\begin{equation}
\max _{\mathbf{A}_v} \operatorname{Tr}\left(\mathbf{A}_v^{\top} \mathbf{M}_v\right), 
\text { s.t. } \mathbf{A}_v^{\top} \mathbf{A}_v=\mathbf{I}_m,
\end{equation}
where $\mathbf{M}_v=\mathbf{X}_v \mathbf{H}_v \mathbf{H}_v^\top \mathbf{Z}_v^\top =\left(\mathbf{X}_v \odot \mathbf{R}_v\right) \mathbf{Z}_v^{\top}$. According to \cite{wang2019multi}, the optimal solution for $\mathbf{A}_v$ is $\mathbf{U}_m \mathbf{V}_m^{\top}$, where $\mathbf{U}_m$ and $\mathbf{V}_m$ represent the matrices which comprise the first $m$ left singular vectors and right singular vectors of $\mathbf{M}_v$, correspondingly. Their time overhead is $\mathcal{O}(nmd+m^2d)$ to obtain all the optimal $\left\{\mathbf{A}_v\right\}_{v=1}^V$, where $d = \sum_{v=1}^V d_v$.

\subsubsection{Optimization of Anchor Graphs \texorpdfstring{$\left\{\mathbf{Z}_v\right\}_{v=1}^V$}{}}
When $\left\{\mathbf{A}_v\right\}_{v=1}^V$, $\left\{\mathbf{P}_v\right\}_{v=1}^V$, $\boldsymbol{\gamma}$ and $\mathbf{F}$ are fixed, the optimization of $\left\{\mathbf{Z}_v\right\}_{v=1}^V$ can be written as follows:
\begin{equation}
\label{opt zv}
 \begin{aligned}
 &\min_{\left\{\mathbf{\mathbf{Z}}_v\right\}_{v=1}^V} \sum_{v=1}^V \gamma_v^2\left\|\mathbf{X}_v \mathbf{\mathbf{H}}_v-\mathbf{A}_v \mathbf{Z}_v \mathbf{H}_v\right\|_F^2\\
& \qquad \ \ +\lambda \sum_{v=1}^V\left\|\mathbf{P}_v \mathbf{Z}_v-\mathbf{F} \right\|_F^2 + \mu \sum_{v=1}^V \left\|\mathbf{Z}_v \right\|_F^2, \\
& \quad \text {s.t.} \quad  \mathbf{Z}_v \geq 0, \mathbf{Z}_v^{\top} \mathbf{1}_m=\mathbf{1}_n.
 \end{aligned} 
\end{equation}

By removing the irrelevant items, the Eq.(\ref{opt zv}) can be rewrited as:
\begin{equation}
\begin{aligned}
& \min _{\mathbf{Z}_v} \operatorname{Tr} \left(\mathbf{Z}_v^{\top} \mathbf{Z}_v\left(\gamma_v^2 \mathbf{H}_v \mathbf{H}_v^{\top}+ \left(\lambda +\mu \right)\mathbf{I}\right)\right)\\
& \quad -2 \operatorname{Tr} \left(\mathbf{Z}_{v}^\top \left(\gamma_v^2 \mathbf{A}_v^\top \mathbf{X}_v \mathbf{H}_v \mathbf{H}_v^{\top} +\lambda \mathbf{P}_v^{\top} \mathbf{F}\right)\right) \\
& \text {s.t.} \quad \mathbf{Z}_v \geq 0, \quad \mathbf{Z}_v^{\top} \mathbf{1}_m=\mathbf{1}_n,
\end{aligned}
\end{equation}
with $\mathbf{X}_v \mathbf{H}_v \mathbf{H}_v{ }^{\top}=\mathbf{X}_v \odot \mathbf{R}_v$. Denoting $\mathbf{z}_j^{(v)}$ as the $j$-th column vector of $\mathbf{Z}_v$, we have

\begin{equation}
\label{opt zv3}
\min_{\mathbf{z}_j^{(v)}} \frac{1}{2}\left\|\mathbf{z}_j^{(v)}-\mathbf{f}_j^{(v)}\right\|_F^2, \quad \text{s.t.} \mathbf{z}_j^{(v)} \geq 0,  \mathbf{z}_j^{(v)\top} \mathbf{1}_m=1, 
\end{equation}
where $\mathbf{f}_{i j}^{(v)} = \frac{\gamma_v^2 \left[\mathbf{A}_v^\top \left( \mathbf{X}_v \odot \mathbf{R}_v\right) \right]_{ij} + \lambda \left[\mathbf{P}_v^{\top} \mathbf{F}\right]_{ij} }{\gamma_v^2 r_j^{(v)}+\lambda + \mu}$, $\left[\mathbf{P}_v^{\top} \mathbf{F}\right]_{ij}$ denotes the element of the $i$-th row and $j$-th column of $\mathbf{P}_v^{\top} \mathbf{F}$.

We write the Lagrangian function of Eq.(\ref{opt zv3}) as
$$
\small
\mathcal{L}\left(\mathbf{z}_j^{(v)}, \alpha_j, \boldsymbol{\eta}_j\right)=\frac{1}{2}\left\|\mathbf{z}_j^{(v)}-\mathbf{f}_j^{(v)}\right\|_F^2-\alpha_j\left(\mathbf{z}_j^{(v)\top} \mathbf{1}_m -1\right)-\boldsymbol{\eta}_j^{\top} \mathbf{z}_j^{(v)},
$$
where $\alpha_j$ and $\boldsymbol{\eta}_j$ represent the respective Lagrange multipliers. Their Kahn-Kuhn-Tucker (KKT) conditions can write as
$$
\left\{\begin{array}{l}
\mathbf{z}_j^{(v)}-\mathbf{f}_j^{(v)}-\alpha_j \mathbf{1}_m-\boldsymbol{\eta}_j=0, \\
\boldsymbol{\eta}_j \odot \mathbf{z}_j^{(v)}= \mathbf{0}.
\end{array}\right.
$$

Together with $\mathbf{z}_j^{(v)\top} \mathbf{1}_m=1$, we can derive the equation below:
$$
\mathbf{z}_j^{(v)}=\max \left(\mathbf{f}_j^{(v)}+\alpha_j \mathbf{1}_m,0\right),
$$
where $\alpha_j$ could be addressed by Newton's method effectively. The time complexity of optimizing $\left\{\mathbf{Z}_v\right\}_{v=1}^V$ is $\mathcal{O}(nmd)$.

\subsubsection{Optimization of Consistent Representation \texorpdfstring{$\mathbf{F}$}{}}
When $\left\{\mathbf{A}_v\right\}_{v=1}^V$, $\left\{\mathbf{P}_v\right\}_{v=1}^V$,  $\left\{\mathbf{Z}_v\right\}_{v=1}^V$ and $\boldsymbol{\gamma}$ are fixed, the optimization for $\mathbf{Z}$ can be written as follows:
\begin{equation} \label{opt F}
\max _{\mathbf{F}} \operatorname{Tr}\left(\mathbf{F} \mathbf{Q}\right), \quad \text {s.t.} \mathbf{F} \mathbf{F}^\top=\mathbf{I}_m,
\end{equation}
where $\mathbf{Q}= \sum_{v=1}^V \mathbf{Z}_v^\top \mathbf{P}_v^{\top}$, the optimal solution of $\mathbf{F}$ is $\boldsymbol{\Psi}_m \boldsymbol{\Sigma}_m ^{\top}$, where $\boldsymbol{\Sigma}_m$ and $\boldsymbol{\Psi}_m$ indicate the matrices which comprise the first $m$ left singular vectors and the first $m$ right singular vectors of $\mathbf{W}_v$, correspondingly. It costs  $\mathcal{O}(nm^2V)$ time.

\begin{algorithm}[t]
	\renewcommand{\algorithmicrequire}{\textbf{Input:}}
	\renewcommand{\algorithmicensure}{\textbf{Output:}}
	\caption{Scalable Incomplete Multi-View Clustering with Structure Alignment (SIMVC-SA)}
	\label{my alg}
 \small
	\begin{algorithmic}[1]
		\REQUIRE $v$ views incomplete dataset $\left\{\mathbf{X}_v\right\}_{v=1}^{V}$, the missing index  $\left\{\mathbf{H}_v\right\}_{v=1}^{V}$, and  the number of cluster $k$.
		
		\STATE Initialize  $\left\{\mathbf{Z}_v\right\}_{v=1}^{V}$, $\left\{\mathbf{P}_v\right\}_{v=1}^{V}$, $\boldsymbol{\gamma}$. 
         \REPEAT
		
		\STATE Obtain $\left\{\mathbf{A}_v\right\}_{v=1}^{V}$ with Eq. (\ref{opt A}).
		\STATE Obtain $\mathbf{F}$ with Eq. (\ref{opt F}).
		\STATE Obtain $\left\{\mathbf{Z}_v\right\}_{v=1}^{V}$ with Eq. (\ref{opt zv}).
  \STATE Obtain $\left\{\mathbf{P}_v\right\}_{v=1}^{V}$ with Eq. (\ref{opt p}).
  \STATE Obtain $\boldsymbol{\gamma}$ with Eq. (\ref{opt gamma}).
  \UNTIL{converged}
\STATE Obtain $\mathbf{U}$ by performing SVD on $\mathbf{F}$
  \ENSURE Perform k-means on $\mathbf{U}$ to obtain discrete labels.
	\end{algorithmic}  
	\label{algo_whole}
\end{algorithm}

\subsubsection{Optimization of Alignment Matrices \texorpdfstring{$\left\{\mathbf{P}_v\right\}_{v=1}^V$}{}}
When $\left\{\mathbf{A}_v\right\}_{v=1}^V$, $\left\{\mathbf{Z}_v\right\}_{v=1}^V$, $\boldsymbol{\gamma}$ and $\mathbf{F}$ are fixed, the optimization for $\left\{\mathbf{P}_v\right\}_{v=1}^V$ can be written as follows:
\begin{equation} \label{opt p}
\max _{\mathbf{P}_v} \operatorname{Tr}\left(\mathbf{P}_v^{\top} \mathbf{W}_v\right), \quad \text {s.t.} \mathbf{P}_v^{\top} \mathbf{P}_v=\mathbf{I}_m,
\end{equation}
where $\mathbf{W}_v=\mathbf{F} \mathbf{Z}_v^{\top}$, Similar to Eq. \eqref{opt F}, this problem can be efficiently solved by rank-k truncated SVD.

\subsubsection{Optimization of View Weight \texorpdfstring{$\boldsymbol{\gamma}$}{}}
When $\left\{\mathbf{A}_v\right\}_{v=1}^V$, $\left\{\mathbf{P}_v\right\}_{v=1}^V$,  $\left\{\mathbf{Z}_v\right\}_{v=1}^V$ and $\mathbf{Z}$ are fixed, the optimization for $\boldsymbol{\gamma}$ can be written as follows:
\begin{equation}
\label{opt gamma}
\min_{\boldsymbol{\gamma}} \sum_{v=1}^V \gamma_v^2 \tau_v, \quad \text {s.t.} \boldsymbol{\gamma}^{\top} \mathbf{1}_V=1, \boldsymbol{\gamma} \geq 0,
\end{equation}
where $\tau_v=\left\|\mathbf{X}_v \mathbf{H}_v-\mathbf{A}_v \mathbf{Z}_v \mathbf{H}_v\right\|_F^2 $. By Cauchy–Schwarz inequality, the view weight $\boldsymbol{\gamma}$ can be acquired by 
\begin{equation}
\gamma_v=\frac{1/\tau_v}{\sum_{v=1}^V 1/\tau_v}.
\end{equation}

 It consumes $\mathcal{O}(nmd)$ time. Algorithm 1 summarises the entire optimization procedure for addressing Eq.(\ref{my formula}).

\subsection{Discussions}
\subsubsection{Convergence}
As the iterations proceed, five variables of the above optimization procedure will be separately addressed. As each sub-optimization problem reaches the global optimum, the objective value monotonically decreases until the convergence condition is attained \cite{Bezdek2003}. Furthermore, because it is easy to prove that the lower boundary of the objective function is zero, our proposed SIMVC-SA can converge to the local optimum.


\subsubsection{Time Complexity}
The time overhead of SIMVC-SA is composed of five optimization processes, as previously mentioned. The time overhead of updating $\left\{\mathbf{A}_v\right\}_{v=1}^V$ is $\mathcal{O}\left(nmd+m^2d\right)$. When updating $\left\{\mathbf{Z}_v\right\}_{v=1}^V$ and $\boldsymbol{\gamma}$ need $\mathcal{O}\left(nmd\right)$. When analytically obtaining $\left\{\mathbf{P}_v\right\}_{v=1}^V$, it costs $\mathcal{O}((nm^2+m^3)V)$ for all columns. The time overhead of calculating $\mathbf{F}$ is $\mathcal{O}(nm^2V)$. As a result, the total time overhead of the optimization procedure is $\mathcal{O}\left(n\left(md+m^2V\right)+m^3V+m^2d\right)$. Consequently, the computational complexity of SIMVC-SA is $\mathcal{O}(n)$,  which is linearly related to the number of samples.

\begin{table}[t]
\caption{Incompete Multiview Datasets
in our Experments}
\vspace{-1 em}
\label{benchmark_data}
\begin{tabular}{ccccc}
\toprule
Dataset     & Size  & Clusters & Views & Dimensionality    \\ \midrule
ORL         & 400   & 40       & 3     & 4096/3304/6750    \\
ProteinFold & 694   & 27       & 12    & 27/27/.../27/27/27         \\
BDGP        & 2500  & 5        & 3     & 1000/500/250      \\
SUNRGBD     & 10335 & 45       & 2     & 4096/4096         \\
NUSWIDEOBJ  & 30000 & 31       & 5     & 65/226/145/74/129 \\
Cifar10    & 50000 & 10       & 3     & 512/2048/1024     \\
MNIST       & 60000 & 10       & 3     & 342/1024/64       \\ \bottomrule
\end{tabular}
\vspace{-1 em}
\end{table}
\begin{table*}[t]
\fontsize{8}{9.7}\selectfont 
\centering

\caption{Empirical evaluation and comparison of SIMVC-SA with twelve baseline methods on seven benchmark datasets.}
\vspace{-1 em}
\resizebox{\linewidth}{!}{
\renewcommand\arraystretch{1}
\tabcolsep=0.1cm
\scalebox{1.07}{
\begin{tabular}{cccccccccccccc}
\toprule
Methods     & BSV        & MIC        & MKKM-IK    & AWP                               & DAIMC                             & APMC                              & UEAF                              & MKKM-IK-MKC & EEIMVC                            & FLSD                              & V$^3$H                               & FIMVC-VIA                          & Proposed                          \\ \hline
\multicolumn{14}{c}{ACC (\%)}                                                                                                                                                                                                                                                                                                                                                                        \\ \hline
ORL         & 24.32±0.89 & 37.56±1.66 & 59.80±2.44 & 68.69±0.00                        & 68.03±2.32                        & 65.58±1.91                        & 60.25±2.50                        & 64.95±2.62  & 73.24±2.54                        & 48.09±1.85                        & 67.03±1.45                        & {\color[HTML]{FF0000} 76.36±2.79} & {\color[HTML]{0070C0} 76.06±2.36} \\
ProteinFold & 22.25±0.53 & 15.99±0.78 & 26.03±1.06 & 28.00±0.00                        & 28.65±1.65                        & N/A                               & {\color[HTML]{0070C0} 28.72±1.53} & 17.99±0.83  & 27.75±1.67                        & 25.98±1.33                        & 17.33±0.48                        & 28.15±1.26                        & {\color[HTML]{FF0000} 30.17±1.23} \\
BDGP        & 34.96±1.06 & 25.37±0.61 & 32.17±0.24 & 23.62±0.00                        & 28.12±0.01                        & 28.12±0.01                        & {\color[HTML]{0070C0} 44.88±0.02} & 40.77±0.20  & 44.00±0.05                        & 42.96±0.03                        & 43.63±0.75                        & 39.84±0.16                        & {\color[HTML]{FF0000} 48.11±0.21} \\
SUNRGBD     & 6.14±0.08  & 14.61±0.54 & 11.35±0.31 & 17.01±0.00                        & 17.03±0.65                        & {\color[HTML]{FF0000} 17.34±0.57} & 15.35±0.41                        & 16.81±0.49  & 16.74±0.49                        & 14.42±0.34                        & N/A                               & 16.88±0.48                        & {\color[HTML]{0070C0} 17.18±0.48} \\
NUSWIDEOBJ  & 12.05±0.03 & N/A        & N/A        & N/A                               & {\color[HTML]{0070C0} 13.79±0.37} & N/A                               & N/A                               & N/A         & 12.73±0.16                        & N/A                               & N/A                               & 12.96±0.12                        & {\color[HTML]{FF0000} 15.40±0.32} \\
Cifar10     & N/A        & N/A        & N/A        & N/A                               & 95.81±0.45                        & N/A                               & N/A                               & N/A         & N/A                               & N/A                               & N/A                               & {\color[HTML]{0070C0} 96.16±0.00} & {\color[HTML]{FF0000} 96.32±0.22} \\
MNIST       & N/A        & N/A        & N/A        & N/A                               & 97.57±0.31                        & N/A                               & N/A                               & N/A         & N/A                               & N/A                               & N/A                               & {\color[HTML]{0070C0} 98.16±0.01} & {\color[HTML]{FF0000} 98.40±0.04} \\ \hline
\multicolumn{14}{c}{NMI (\%)}                                                                                                                                                                                                                                                                                                                                                                        \\ \hline
ORL         & 48.49±0.90 & 56.44±1.00 & 75.95±1.33 & 83.79±0.00                        & 82.89±1.06                        & 80.20±0.82                        & 76.16±1.25                        & 79.76±1.41  & 85.37±1.32                        & 67.91±1.28                        & 81.05±0.61                        & {\color[HTML]{FF0000} 88.08±1.31} & {\color[HTML]{0070C0} 87.53±1.25} \\
ProteinFold & 27.60±0.59 & 16.64±1.02 & 33.70±0.84 & 36.17±0.00                        & {\color[HTML]{0070C0}37.67±1.08}                        & N/A                               & {\color[HTML]{FF0000} 38.18±0.88} & 24.88±0.84  & 36.03±0.98                        & 35.66±0.79                        & 22.75±0.53                        & 36.22±0.96                        & {\color[HTML]{0070C0} 37.72±0.98} \\
BDGP        & 12.88±0.94 & 4.47±0.70  & 7.41±0.16  & 4.68±0.00                         & 8.68±0.01                         & 8.68±0.01                         & 23.55±0.04                        & 16.35±0.13  & 19.91±0.09                        & 18.95±0.06                        & {\color[HTML]{FF0000} 24.15±0.40} & 15.11±0.10                        & {\color[HTML]{0070C0} 23.67±0.14} \\
SUNRGBD     & 3.27±0.08  & 21.27±0.35 & 15.27±0.25 & {\color[HTML]{FF0000}23.60±0.00}                        & 21.53±0.43                        &  22.46±0.25 & 21.72±0.22                        & 20.48±0.28  & 20.84±0.28                        & 20.82±0.17                        & N/A                               & 21.48±0.33                        & {\color[HTML]{0070C0} 22.52±0.23} \\
NUSWIDEOBJ  & 2.68±0.03  & N/A        & N/A        & N/A                               & {\color[HTML]{0070C0} 11.70±0.36} & N/A                               & N/A                               & N/A         & 10.31±0.16                        & N/A                               & N/A                               & 10.27±0.07                        & {\color[HTML]{FF0000} 11.78±0.11} \\
Cifar10     & N/A        & N/A        & N/A        & N/A                               & 90.47±0.35                        & N/A                               & N/A                               & N/A         & N/A                               & N/A                               & N/A                               & {\color[HTML]{0070C0} 91.18±0.00} & {\color[HTML]{FF0000} 91.21±0.18} \\
MNIST       & N/A        & N/A        & N/A        & N/A                               & 93.89±0.53                        & N/A                               & N/A                               & N/A         & N/A                               & N/A                               & N/A                               & {\color[HTML]{FF0000} 95.76±0.02} & {\color[HTML]{0070C0} 95.58±0.00} \\ \hline
\multicolumn{14}{c}{Purity (\%)}                                                                                                                                                                                                                                                                                                                                                                     \\ \hline
ORL         & 26.80±0.92 & 40.81±1.40 & 62.79±2.11 & 70.42±0.00                        & 71.82±1.79                        & 69.24±1.48                        & 63.90±1.90                        & 67.68±2.34  & 76.09±2.19                        & 50.88±1.72                        & 70.22±1.09                        & {\color[HTML]{0070C0} 78.59±2.29} & {\color[HTML]{FF0000} 78.79±2.08} \\
ProteinFold & 25.89±0.60 & 19.78±0.84 & 30.91±1.04 & 31.97±0.00                        & 34.99±1.54                        & N/A                               & {\color[HTML]{0070C0} 35.47±1.16} & 22.73±0.87  & 33.13±1.30                        & 32.82±0.97                        & 22.24±0.51                        & 33.67±1.12                        & {\color[HTML]{FF0000} 35.97±1.14} \\
BDGP        & 36.75±0.89 & 25.67±0.59 & 33.42±0.18 & 24.02±0.00                        & 28.46±0.01                        & 28.46±0.01                        & 45.92±0.01                        & 41.10±0.13  & {\color[HTML]{0070C0} 46.40±0.05} & 44.41±0.03                        & 45.38±0.50                        & 40.12±0.15                        & {\color[HTML]{FF0000} 49.05±0.19} \\
SUNRGBD     & 13.06±0.16 & 32.36±0.59 & 27.06±0.45 & {\color[HTML]{FF0000} 37.56±0.00} & 34.39±0.59                        & 33.19±0.47                        & 33.37±0.48                        & 32.92±0.50  & 33.58±0.48                        & 32.41±0.39                        & N/A                               & 34.34±0.61                        & {\color[HTML]{0070C0} 34.51±0.51} \\
NUSWIDEOBJ  & 13.72±0.04 & N/A        & N/A        & N/A                               & {\color[HTML]{0070C0} 23.41±0.63} & N/A                               & N/A                               & N/A         & 21.83±0.21                        & N/A                               & N/A                               & 21.97±0.12                        & {\color[HTML]{FF0000} 23.71±0.27} \\
Cifar10     & N/A        & N/A        & N/A        & N/A                               & 95.81±0.45                        & N/A                               & N/A                               & N/A         & N/A                               & N/A                               & N/A                               & {\color[HTML]{0070C0} 96.23±0.00} & {\color[HTML]{FF0000} 96.29±0.25} \\
MNIST       & N/A        & N/A        & N/A        & N/A                               & 97.57±0.31                        & N/A                               & N/A                               & N/A         & N/A                               & N/A                               & N/A                               & {\color[HTML]{0070C0} 98.24±0.02} & {\color[HTML]{FF0000} 98.42±0.05} \\ \hline
\multicolumn{14}{c}{Fscore (\%)}                                                                                                                                                                                                                                                                                                                                                                     \\ \hline
ORL         & 9.01±0.69  & 17.30±1.18 & 46.32±2.50 & 58.73±0.00                        & 56.84±2.87                        & 50.70±2.55                        & 42.53±2.74                        & 53.30±2.91  & 63.67±2.85                        & 31.17±2.00                        & 54.27±1.40                        & {\color[HTML]{FF0000} 68.25±3.25} & {\color[HTML]{0070C0} 67.62±2.85} \\
ProteinFold & 12.30±0.08 & 10.35±0.38 & 14.35±0.81 & 12.38±0.00                        & {\color[HTML]{FF0000} 16.97±1.09} & N/A                               & 16.31±1.22                        & 8.92±0.48   & 15.62±1.19                        & 14.57±1.02                        & 10.33±0.13                        & 15.53±1.00                        & {\color[HTML]{0070C0} 16.81±1.04} \\
BDGP        & 28.76±0.61 & 29.88±0.06 & 25.25±0.08 & 32.53±0.00                        & 31.21±0.00                        & 31.21±0.00                        & 33.69±0.03                        & 30.15±0.10  & 32.93±0.04                        & 34.29±0.01                        & {\color[HTML]{0070C0} 35.31±0.24} & 31.65±0.10                        & {\color[HTML]{FF0000} 35.40±0.12} \\
SUNRGBD     & 6.89±0.01  & 9.45±0.24  & 7.10±0.14  & 11.58±0.00                        & 10.71±0.35                        & 10.96±0.25                        & 10.20±0.16                        & 10.06±0.21  & 10.20±0.20                        & {\color[HTML]{FF0000} 12.09±0.00} & N/A                               & 10.33±0.24                        & {\color[HTML]{0070C0} 11.74±0.10} \\
NUSWIDEOBJ  & 10.95±0.00 & N/A        & N/A        & N/A                               & {\color[HTML]{0070C0} 8.58±0.19}  & N/A                               & N/A                               & N/A         & 7.81±0.08                         & N/A                               & N/A                               & 7.87±0.07                         & {\color[HTML]{FF0000} 11.77±0.12} \\
Cifar10     & N/A        & N/A        & N/A        & N/A                               & 92.16±0.68                        & N/A                               & N/A                               & N/A         & N/A                               & N/A                               & N/A                               & {\color[HTML]{0070C0} 92.79±0.00} & {\color[HTML]{FF0000} 92.90±0.19} \\
MNIST       & N/A        & N/A        & N/A        & N/A                               & 95.28±0.57                        & N/A                               & N/A                               & N/A         & N/A                               & N/A                               & N/A                               & {\color[HTML]{0070C0} 96.66±0.00} & {\color[HTML]{FF0000} 96.92±0.00}\\
\bottomrule
\end{tabular}}}
\label{results}
\setlength{\abovecaptionskip}{-2pt}
\end{table*}

\section{Experiment}

\subsection{Datasets}
Seven wide-used datasets are adopted to evaluate the effectiveness of the proposed algorithm, including ORL, ProteinFold, BDGP, SUNRGBD, NUSWIDEOBJ, Cifar10, and MNIST. The elaborate information of these datasets is listed in Tab. \ref{benchmark_data}. For the above datasets, we remove samples randomly on each view to obtain its incomplete version. Specifically, according to \cite{li2022high}, with the principle that each sample appears in at least one view, we generate incomplete datasets at 0.1 intervals from 0.1 to 0.9.

\begin{figure*}[t]
\centering
\includegraphics[width=0.9\linewidth]{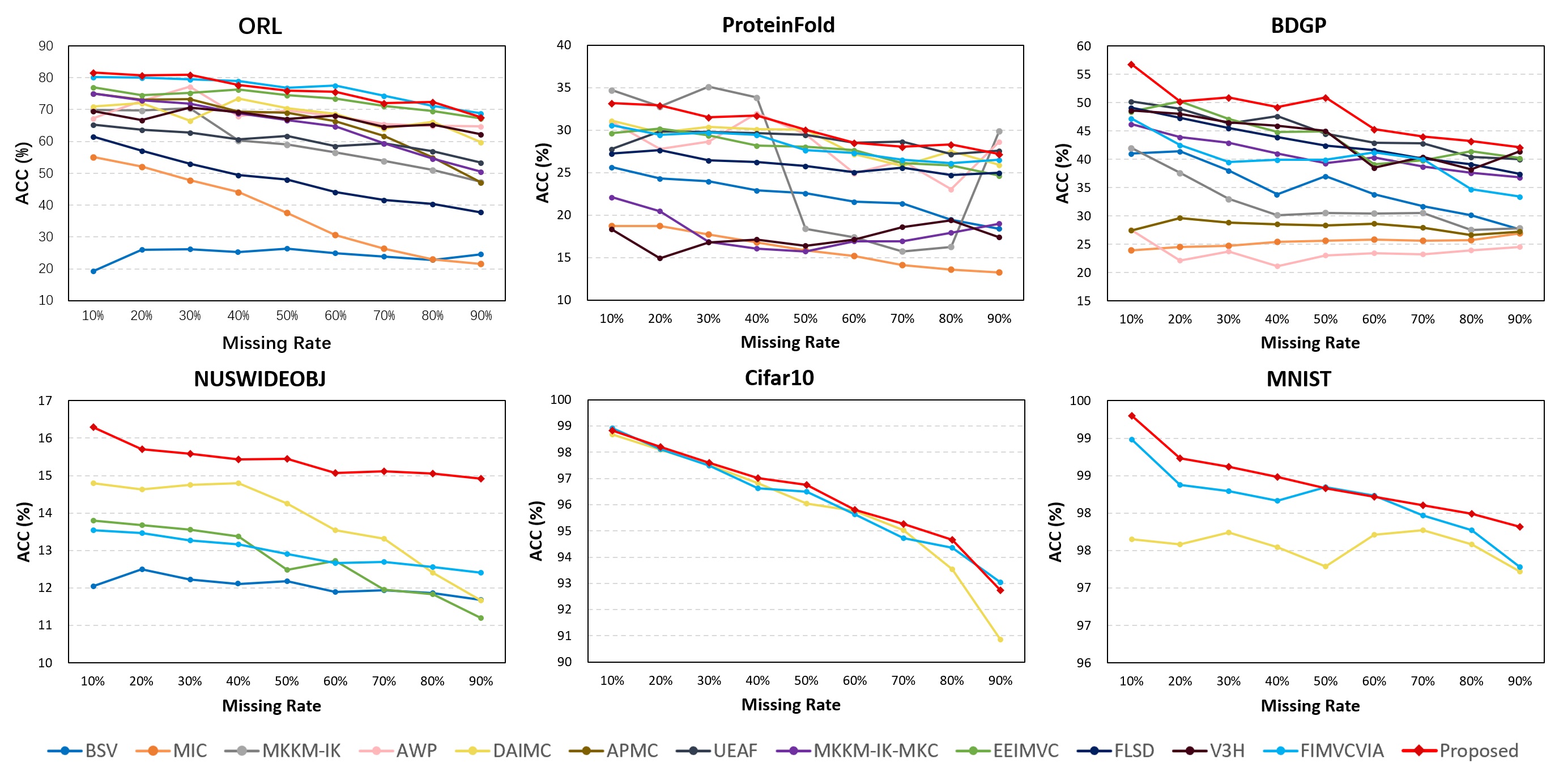}
\vspace{-1 em}
\caption{Clustering performance of SIMVC-SA on benchmark datasets with different missing ratio.}

\label{par_result}
\end{figure*}

\begin{figure*}[t]
\centering
\includegraphics[width=0.85\linewidth]{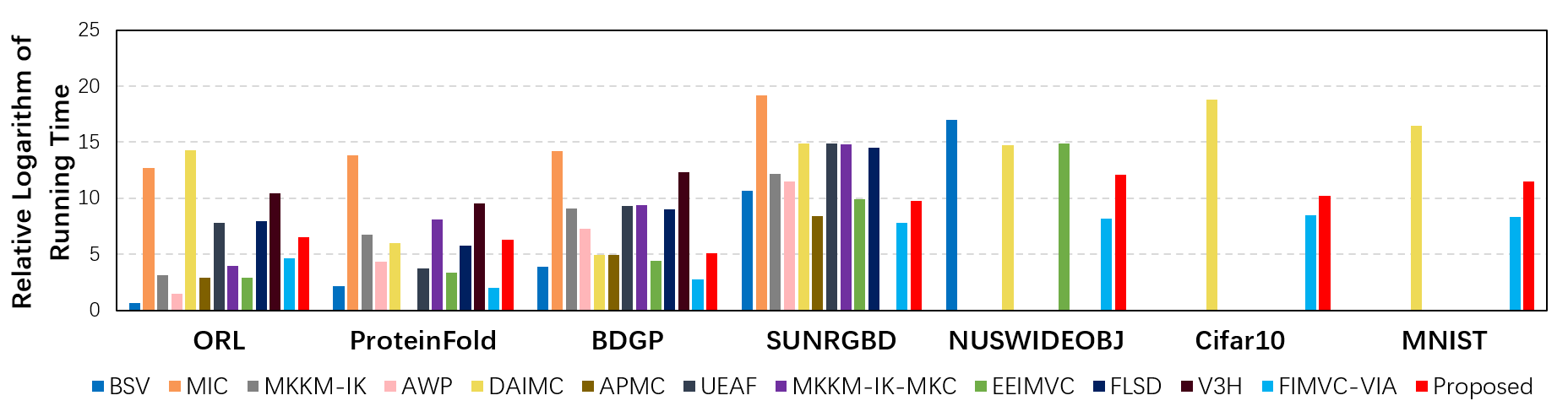}
\vspace{-1 em}
\caption{Time Comparison of Different IMVC Methods on Seven Incomplete Datasets}
\label{time}
\setlength{\abovecaptionskip}{-2pt}
\end{figure*}
\begin{figure*}
\centering
\vspace{-1 em}
\includegraphics[width=0.93\linewidth]{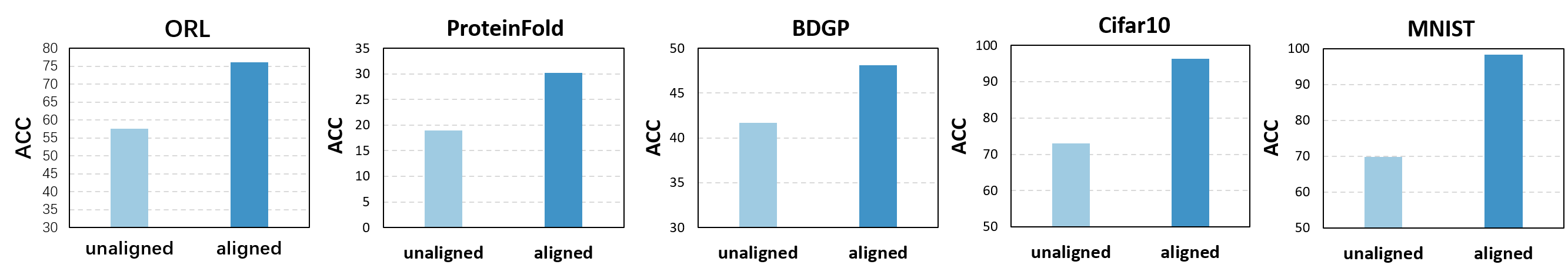}
\caption{The ablation study of our structural alignment strategy on five benchmark datasets. "Unaligned" indicates without our structural alignment strategy.}
\label{unalign}
\setlength{\abovecaptionskip}{-2pt}
\end{figure*}

\begin{figure}
\centering
\setlength{\abovecaptionskip}{-1pt}
\includegraphics[width=1\linewidth]{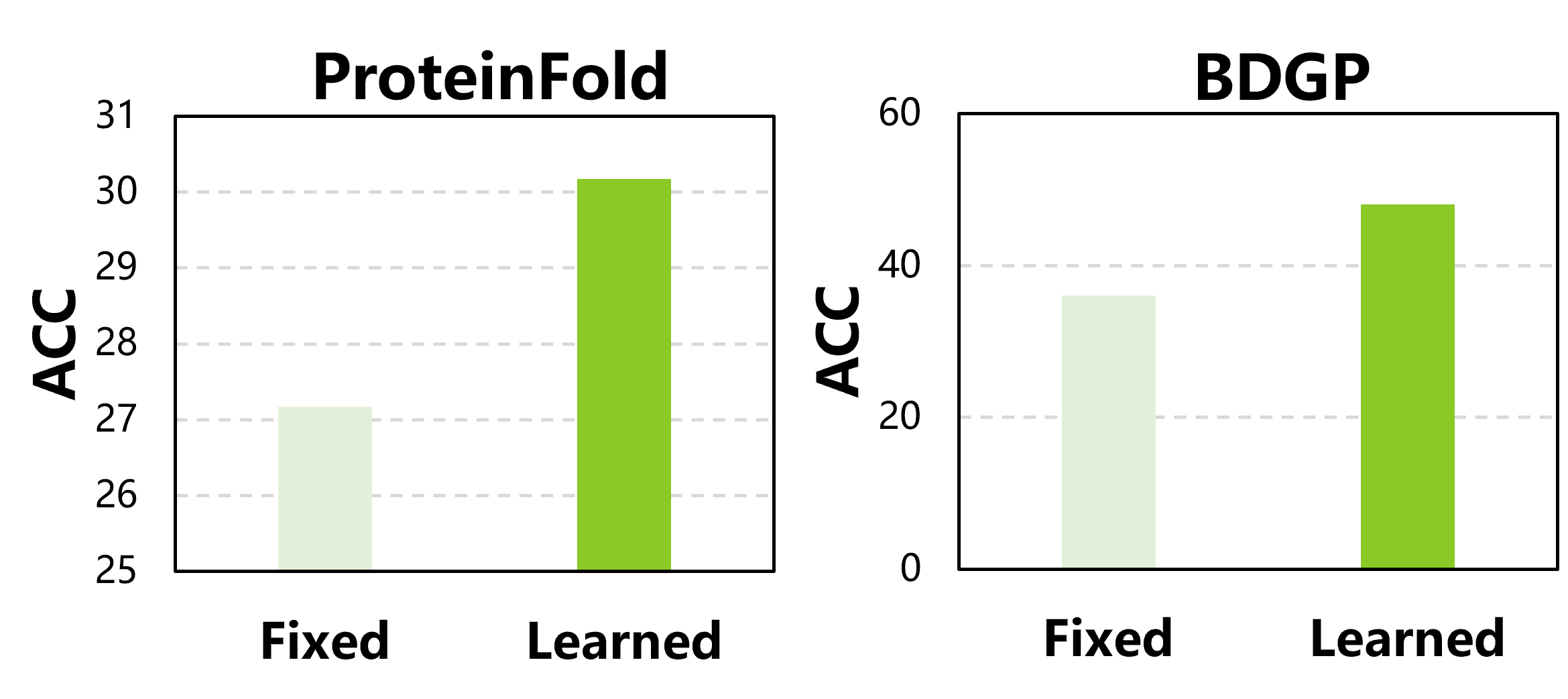}
\caption{The ablation study of our anchor learning strategy. }
\label{obection}
\vspace{-1 em}
\end{figure}

\subsection{Compared Methods and Setting}
Along with our proposed SIMVC-SA,  we run twelve state-of-the-art incomplete multi-view clustering methods for comparison, including Best Single View (BSV) \cite{ng2002spectral}, MIC Views Clustering via Weighted NMF With $\ell_{2,1}$ Regularization (MIC) \cite{shao2015multiple}, Multiple Kernel k-Means With Incomplete Kernels (MKKM-IK) \cite{liu2017multiple}, Multiview Clustering via Adaptively Weighted Procrustes (AWP) \cite{nie2018multiview}, Doubly Aligned Incomplete Multi-view Clustering (DAIMC) \cite{hu2019doubly}, Anchor-based partial multi-view clustering (APMC) \cite{guo2019anchors}, Unified Embedding Alignment With Missing Views Inferring for Incomplete Multiview Clustering (UEAF) \cite{wen2019unified}, Multiple Kernel k k-Means With Incomplete Kernels and Multiple Kernel Clustering (MKKM-IK-MKC) \cite{liu2020multiple}, Efficient and Effective Regularized Incomplete Multiview Clustering (EEIMVC) \cite{liu2020efficient}, Generalized IMVC With Flexible locality Structure Diffusion (FLSD) \cite{wen2020generalized}, View Variation and View Heredity for
Incomplete Multiview Clustering (V$^3$H) \cite{fang2020v} and Fast Incomplete Multi-View Clustering With
View Independent Anchors (FIMVC-VIA) \cite{liu2022fast}. 

For all the algorithms mentioned above, we set their parameters
as their recommended range. In the proposed method, we adjusted $\lambda$ to $[10^{-4},10^{-2},1,10^2,10^4]$, $\mu$ to
$[0, 10^{-4},10^{-2},1,10^2,10^4]$, and the anchor numbers of [k, 2k, 5k] using a mesh search scheme. In addition, we repeated each experiment 10 cycles to calculate the average performance and standard bias. To assess the clustering performance, we employ four well-used criteria consisting of accuracy (ACC), normalized mutual information (NMI), Purity, and Fscore. All experiments were conducted on a desktop computer with Intel core i9-10900X CPU and 64G RAM, MATLAB 2020b (64-bit).

\vspace{-10 pt}

\subsection{Experimental Results}

Tab. \ref{results} reports the clustering results on seven benchmark datasets. The best results are marked in red, while the second-best results are marked in blue. "N/A" indicates the unavailable results due to time-out or out-of-memory errors. Besides, we compare the ACC of all methods with different missing rates in Fig. \ref{par_result}. 
According to the results, we have the following conclusions:

\begin{figure}
\centering
\setlength{\abovecaptionskip}{-2pt}
\includegraphics[width=1\linewidth]{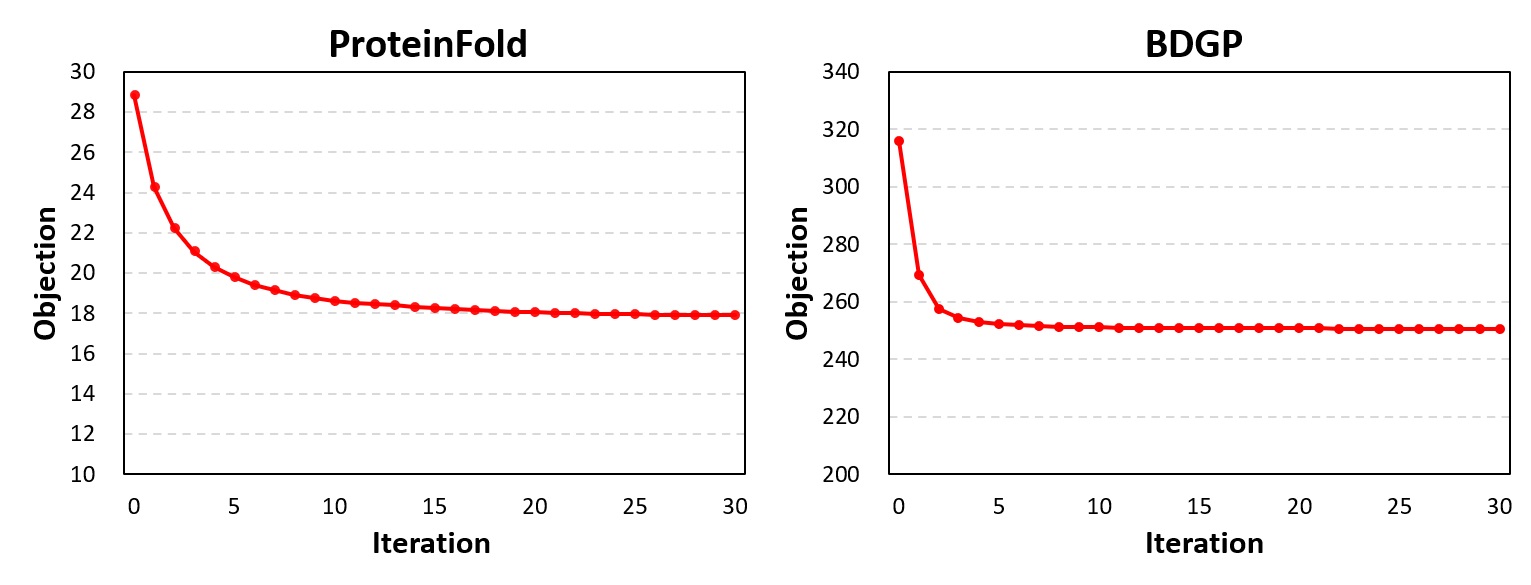}
\caption{Objectives of the proposed method.}
\label{fix}
\vspace{-1 em}
\end{figure}
\begin{enumerate}[(1)]
    \item Compared with existing IMVC methods, our proposed algorithm demonstrates the best performance in most datasets. The recently proposed FIMVC-VIA method shows better performance than other methods, which demonstrates its superiority in incomplete datasets. In terms of ACC, our SIMVC-SA achieves better performance than FIMVC-VIA on the ProteinFold, BDGP, SUNRGBD, NUSWIDEOBJ, and Cifar10 datasets, i.e., 2.02\%, 8.27\%, 0.3\%, 2.44\%, and 0.16\%, which demonstrates the effectiveness of view-specific representation and cross-view alignment strategy. 
    \item Compared to traditional subspace-based IMVC methods, our anchor-based method achieves the best performance in most cases and is applicable to various large-scale datasets.
    \item As shown in Fig. \ref{par_result}, we can observe that most IMVC methods show greater fluctuations in performance with the missing rate rising, while our method is more stable. We conjecture that this is because the alignment of the representations well complements the missing information of different views.
\end{enumerate}

\subsection{Running Time Comparison}
To validate the computational efficiency of the proposed SIMVC-SA, we plot the average running time of each algorithm on seven benchmark datasets in Fig. \ref{time}. The results of some compared algorithms on large-scale datasets are not reported due to memory overflow errors. As shown in the Fig. \ref{time}, we can observe that
\begin{enumerate} [(1)]
    \item Compared to full graph-based clustering methods, the proposed SIMVC-SA significantly reduces run time through the construction of anchor graphs.
    \item Compared to the anchor-based IMVC approach, i.e., FIMVC-VIA, the proposed SIMVC-SA requires more time consumption, mainly due to our view-specific representation and structure alignment strategy, the extra computational complexity increases with the number of views, which is most obvious in NUSWIDEOBJ (5 views). General, the extra time spent is worthwhile since our proposed SIMVC-SA demonstrates its superiority over FIMVC-VIA in most datasets.
\end{enumerate}

\subsection{Ablation Study}

\paragraph{Structural Alignment Strategy}
The structural alignment strategy is the main contribution of this paper. To further demonstrate the effectiveness of this strategy, we present the experimental results of the ablation study in Fig. \ref{unalign}, where "Unaligned" indicates not using our structural alignment strategy. In our experimental setting, we fixed the alignment matrix $\mathbf{P}$ in the optimization process to obtain the final clustering result. The effectiveness of the proposed strategy can be clearly demonstrated in Fig. \ref{unalign}. In terms of ACC, the proposed structural alignment strategy improves the algorithm performance on the ORL, ProteinFold, BDGP, Cifar10, and MNIST datasets by \textbf{18.46\%,11.16\%, 6.43\%, 23.38\%}, and \textbf{28.55}\% respectively, which demonstrates the effectiveness of our strategy.

\paragraph{Anchor Learning Strategy}
We conducted ablation experiments with the proposed anchor learning strategy, as shown in Fig. \ref{fix}. "Fixed" indicates initializing anchors by k-means without updating during the optimization process. Compared to the above methods, our approach significantly improves the clustering performance and avoids the high time expenditure of k-means. 


\begin{figure}[t]
\setlength{\abovecaptionskip}{-2pt}
\includegraphics[width=1\linewidth]{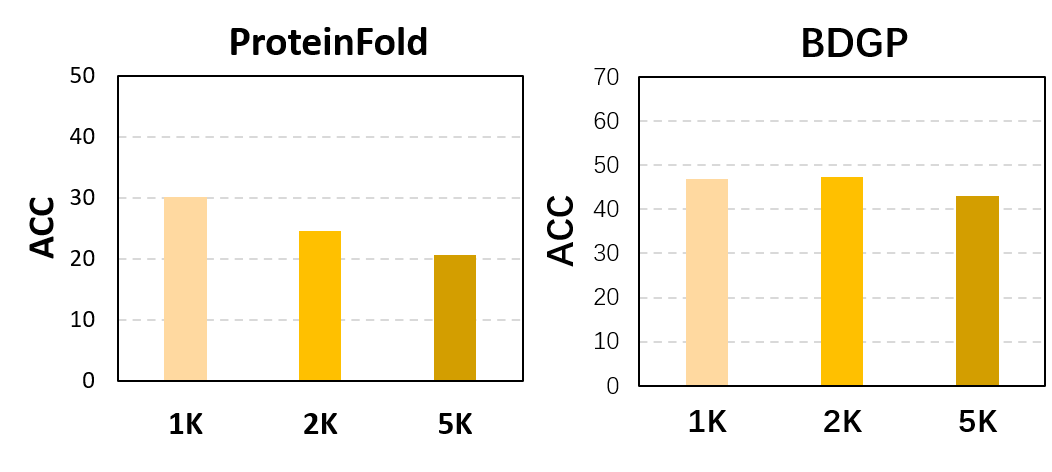}
\caption{Sensitivity analysis of anchor number m  of out method on two benchmark datasets.}
\vspace{-1 em}
\label{anchor}
\end{figure}

\begin{figure}[t]
\centering    
\includegraphics[width=0.99\linewidth]{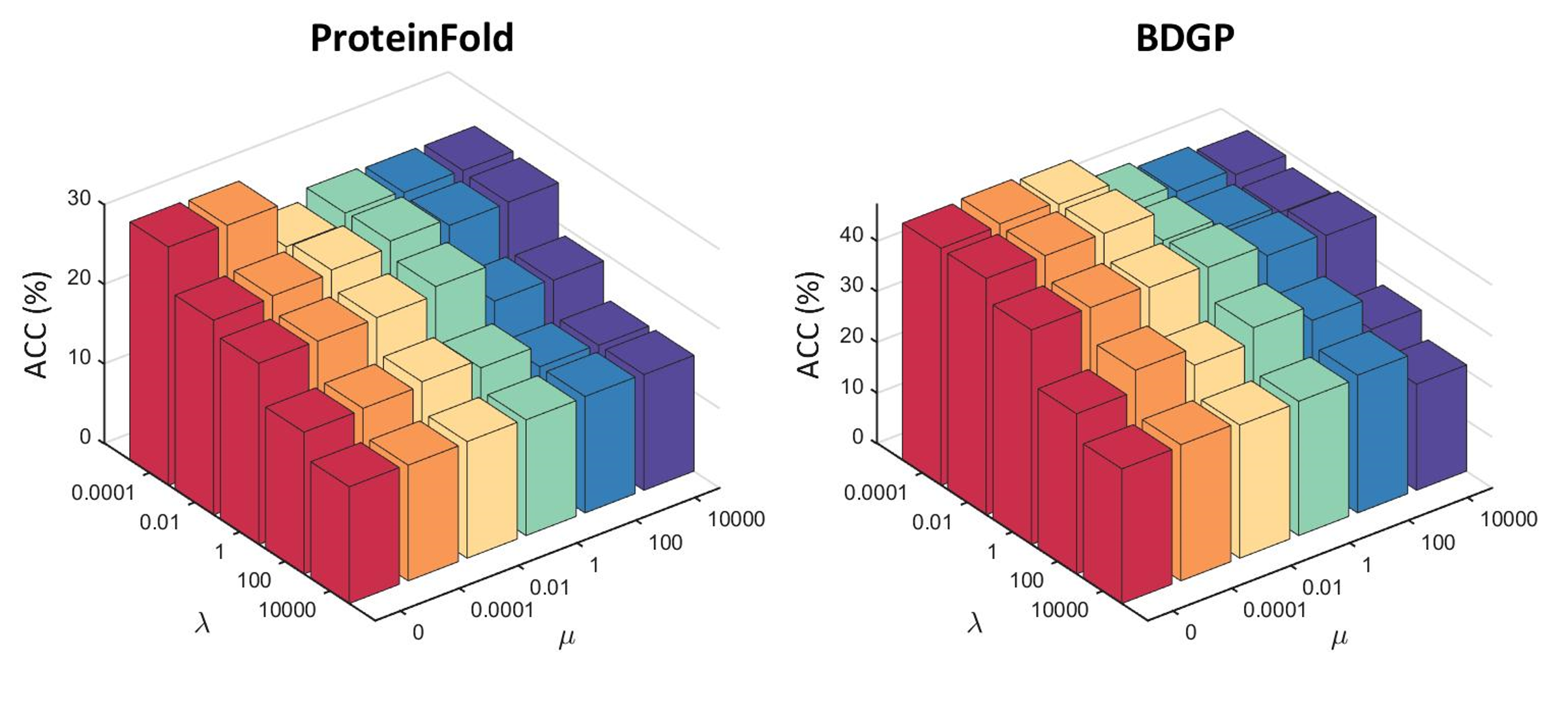}     
\caption{Sensitivity analysis of $\boldsymbol{\lambda}$ and $\boldsymbol{\mu}$ of out method on two benchmark datasets.}     
\label{ablation}    
\vspace{-2 em}
\end{figure}

\subsection{Convergence and Sensitivity}
We conducted several experiments to exhibit the convergence of the proposed SIMVC-SA. As shown in Fig. \ref{obection}, the objective value of our algorithm is monotonically decreasing in each iteration. These results clearly verify the convergence of our proposed algorithm. 

To investigate the sensitivity of SIMVC-SA to the number of anchors m, we investigated how our performance shifts for different numbers of anchors. As shown in Fig. \ref{anchor}, the number of anchors has little effect on the performance of our algorithm. Moreover, two hyperparameters, $\lambda$, and $\mu$, are used in our method, $\lambda$ is the structural alignment parameter, and $\mu$ is the coefficient of the sparsity regularization term. As is shown in Fig. \ref{ablation}, we conducted comparative experiments to indicate the effect of these two parameters on performance.

\section{Conclusion}
In this paper, we propose a novel incomplete anchor graph learning framework termed Scalable Incomplete Multi-View Clustering with Structure Alignment (SIMVC-SA). Specially, we construct the incomplete anchor graph on each view in terms of the unaligned anchor. Besides, a novel structure alignment module is proposed to refine the cross-view anchor correspondence. Meanwhile, the anchor graph construction and alignment are jointly optimized in our unified framework to enhance clustering quality. Through anchor graph construction instead of full graphs, the time and space complexity of our proposed SIMVC-SA is proven to be linearly related to the number of samples. Extensive experiments on seven incomplete benchmark datasets demonstrate the effectiveness and efficiency of our proposed method. In the future, we will explore more flexible alignment strategies. For example, how to align the anchor with different numbers.

\section{ACKNOWLEDGMENTS}
This work was supported by the National Key R\&D Program of
China (no. 2020AAA0107100) and the National Natural Science Foundation of China (project no. 62325604, 62276271).


\bibliographystyle{ACM-Reference-Format}
\balance
\bibliography{sample-base}


\begin{thebibliography}{86}


\ifx \showCODEN    \undefined \def \showCODEN     #1{\unskip}     \fi
\ifx \showDOI      \undefined \def \showDOI       #1{#1}\fi
\ifx \showISBNx    \undefined \def \showISBNx     #1{\unskip}     \fi
\ifx \showISBNxiii \undefined \def \showISBNxiii  #1{\unskip}     \fi
\ifx \showISSN     \undefined \def \showISSN      #1{\unskip}     \fi
\ifx \showLCCN     \undefined \def \showLCCN      #1{\unskip}     \fi
\ifx \shownote     \undefined \def \shownote      #1{#1}          \fi
\ifx \showarticletitle \undefined \def \showarticletitle #1{#1}   \fi
\ifx \showURL      \undefined \def \showURL       {\relax}        \fi
\providecommand\bibfield[2]{#2}
\providecommand\bibinfo[2]{#2}
\providecommand\natexlab[1]{#1}
\providecommand\showeprint[2][]{arXiv:#2}

\bibitem[Bezdek and Hathaway(2003)]%
        {Bezdek2003}
\bibfield{author}{\bibinfo{person}{James~C. Bezdek} {and}
  \bibinfo{person}{Richard~J. Hathaway}.} \bibinfo{year}{2003}\natexlab{}.
\newblock \showarticletitle{Convergence of Alternating Optimization}.
\newblock \bibinfo{journal}{\emph{Neural, Parallel Sci. Comput.}}
  (\bibinfo{year}{2003}).
\newblock


\bibitem[Cai et~al\mbox{.}(2013)]%
        {cai2013multi}
\bibfield{author}{\bibinfo{person}{Xiao Cai}, \bibinfo{person}{Feiping Nie},
  {and} \bibinfo{person}{Heng Huang}.} \bibinfo{year}{2013}\natexlab{}.
\newblock \showarticletitle{Multi-view k-means clustering on big data}. In
  \bibinfo{booktitle}{\emph{Twenty-Third International Joint conference on
  artificial intelligence}}.
\newblock


\bibitem[Fang et~al\mbox{.}(2020)]%
        {fang2020v}
\bibfield{author}{\bibinfo{person}{Xiang Fang}, \bibinfo{person}{Yuchong Hu},
  \bibinfo{person}{Pan Zhou}, {and} \bibinfo{person}{Dapeng~Oliver Wu}.}
  \bibinfo{year}{2020}\natexlab{}.
\newblock \showarticletitle{V $^3$ H: View Variation and View Heredity for
  Incomplete Multiview Clustering}.
\newblock \bibinfo{journal}{\emph{IEEE Transactions on Artificial
  Intelligence}} (\bibinfo{year}{2020}).
\newblock


\bibitem[Gao et~al\mbox{.}(2015)]%
        {gao2015multi}
\bibfield{author}{\bibinfo{person}{Hongchang Gao}, \bibinfo{person}{Feiping
  Nie}, \bibinfo{person}{Xuelong Li}, {and} \bibinfo{person}{Heng Huang}.}
  \bibinfo{year}{2015}\natexlab{}.
\newblock \showarticletitle{Multi-view subspace clustering}. In
  \bibinfo{booktitle}{\emph{Proceedings of the IEEE international conference on
  computer vision}}. \bibinfo{pages}{4238--4246}.
\newblock


\bibitem[Gao et~al\mbox{.}(2023)]%
        {gao2023softclip}
\bibfield{author}{\bibinfo{person}{Yuting Gao}, \bibinfo{person}{Jinfeng Liu},
  \bibinfo{person}{Zihan Xu}, \bibinfo{person}{Tong Wu}, \bibinfo{person}{Wei
  Liu}, \bibinfo{person}{Jie Yang}, \bibinfo{person}{Ke Li}, {and}
  \bibinfo{person}{Xing Sun}.} \bibinfo{year}{2023}\natexlab{}.
\newblock \showarticletitle{SoftCLIP: Softer Cross-modal Alignment Makes CLIP
  Stronger}.
\newblock \bibinfo{journal}{\emph{arXiv preprint arXiv:2303.17561}}
  (\bibinfo{year}{2023}).
\newblock


\bibitem[Gong et~al\mbox{.}(2022)]%
        {gong2022gromov}
\bibfield{author}{\bibinfo{person}{Fengjiao Gong}, \bibinfo{person}{Yuzhou
  Nie}, {and} \bibinfo{person}{Hongteng Xu}.} \bibinfo{year}{2022}\natexlab{}.
\newblock \showarticletitle{Gromov-Wasserstein multi-modal alignment and
  clustering}. In \bibinfo{booktitle}{\emph{Proceedings of the 31st ACM
  International Conference on Information \& Knowledge Management}}.
  \bibinfo{pages}{603--613}.
\newblock


\bibitem[Guo and Ye(2019)]%
        {guo2019anchors}
\bibfield{author}{\bibinfo{person}{Jun Guo} {and} \bibinfo{person}{Jiahui Ye}.}
  \bibinfo{year}{2019}\natexlab{}.
\newblock \showarticletitle{Anchors bring ease: An embarrassingly simple
  approach to partial multi-view clustering}. In
  \bibinfo{booktitle}{\emph{Proc. of AAAI}}.
\newblock


\bibitem[Han et~al\mbox{.}(2022)]%
        {han2022incomplete}
\bibfield{author}{\bibinfo{person}{Xuemei Han}, \bibinfo{person}{Zhenwen Ren},
  \bibinfo{person}{Chuanyun Zou}, {and} \bibinfo{person}{Xiaojian You}.}
  \bibinfo{year}{2022}\natexlab{}.
\newblock \showarticletitle{Incomplete multi-view subspace clustering based on
  missing-sample recovering and structural information learning}.
\newblock \bibinfo{journal}{\emph{Expert Systems with Applications}}
  \bibinfo{volume}{208} (\bibinfo{year}{2022}), \bibinfo{pages}{118165}.
\newblock


\bibitem[Hu and Chen(2018)]%
        {hu2019doubly}
\bibfield{author}{\bibinfo{person}{Menglei Hu} {and} \bibinfo{person}{Songcan
  Chen}.} \bibinfo{year}{2018}\natexlab{}.
\newblock \showarticletitle{Doubly Aligned Incomplete Multi-view Clustering}.
  In \bibinfo{booktitle}{\emph{Proc. of IJCAI}}.
\newblock


\bibitem[Hu and Chen(2019)]%
        {hu2019one}
\bibfield{author}{\bibinfo{person}{Menglei Hu} {and} \bibinfo{person}{Songcan
  Chen}.} \bibinfo{year}{2019}\natexlab{}.
\newblock \showarticletitle{One-pass incomplete multi-view clustering}. In
  \bibinfo{booktitle}{\emph{Proc. of AAAI}}.
\newblock


\bibitem[Huang et~al\mbox{.}(2019)]%
        {huang2019auto}
\bibfield{author}{\bibinfo{person}{Shudong Huang}, \bibinfo{person}{Zhao Kang},
  \bibinfo{person}{Ivor~W Tsang}, {and} \bibinfo{person}{Zenglin Xu}.}
  \bibinfo{year}{2019}\natexlab{}.
\newblock \showarticletitle{Auto-weighted multi-view clustering via kernelized
  graph learning}.
\newblock \bibinfo{journal}{\emph{Pattern Recognition}} (\bibinfo{year}{2019}).
\newblock


\bibitem[Huang et~al\mbox{.}(2020)]%
        {huang2020partially}
\bibfield{author}{\bibinfo{person}{Zhenyu Huang}, \bibinfo{person}{Peng Hu},
  \bibinfo{person}{Joey~Tianyi Zhou}, \bibinfo{person}{Jiancheng Lv}, {and}
  \bibinfo{person}{Xi Peng}.} \bibinfo{year}{2020}\natexlab{}.
\newblock \showarticletitle{Partially view-aligned clustering}.
\newblock \bibinfo{journal}{\emph{Advances in Neural Information Processing
  Systems}}  \bibinfo{volume}{33} (\bibinfo{year}{2020}),
  \bibinfo{pages}{2892--2902}.
\newblock


\bibitem[Jin et~al\mbox{.}(2023)]%
        {jin2023deep}
\bibfield{author}{\bibinfo{person}{Jiaqi Jin}, \bibinfo{person}{Siwei Wang},
  \bibinfo{person}{Zhibin Dong}, \bibinfo{person}{Xinwang Liu}, {and}
  \bibinfo{person}{En Zhu}.} \bibinfo{year}{2023}\natexlab{}.
\newblock \showarticletitle{Deep Incomplete Multi-view Clustering with
  Cross-view Partial Sample and Prototype Alignment}.
\newblock \bibinfo{journal}{\emph{arXiv preprint arXiv:2303.15689}}
  (\bibinfo{year}{2023}).
\newblock


\bibitem[Kang et~al\mbox{.}(2020a)]%
        {kang2020partition}
\bibfield{author}{\bibinfo{person}{Zhao Kang}, \bibinfo{person}{Xinjia Zhao},
  \bibinfo{person}{Chong Peng}, \bibinfo{person}{Hongyuan Zhu},
  \bibinfo{person}{Joey~Tianyi Zhou}, \bibinfo{person}{Xi Peng},
  \bibinfo{person}{Wenyu Chen}, {and} \bibinfo{person}{Zenglin Xu}.}
  \bibinfo{year}{2020}\natexlab{a}.
\newblock \showarticletitle{Partition level multiview subspace clustering}.
\newblock \bibinfo{journal}{\emph{Neural Networks}} (\bibinfo{year}{2020}).
\newblock


\bibitem[Kang et~al\mbox{.}(2020b)]%
        {kang2020large}
\bibfield{author}{\bibinfo{person}{Zhao Kang}, \bibinfo{person}{Wangtao Zhou},
  \bibinfo{person}{Zhitong Zhao}, \bibinfo{person}{Junming Shao},
  \bibinfo{person}{Meng Han}, {and} \bibinfo{person}{Zenglin Xu}.}
  \bibinfo{year}{2020}\natexlab{b}.
\newblock \showarticletitle{Large-scale multi-view subspace clustering in
  linear time}. In \bibinfo{booktitle}{\emph{Proceedings of the AAAI Conference
  on Artificial Intelligence}}, Vol.~\bibinfo{volume}{34}.
  \bibinfo{pages}{4412--4419}.
\newblock


\bibitem[Khan and Maji(2019)]%
        {khan2019approximate}
\bibfield{author}{\bibinfo{person}{Aparajita Khan} {and}
  \bibinfo{person}{Pradipta Maji}.} \bibinfo{year}{2019}\natexlab{}.
\newblock \showarticletitle{Approximate graph Laplacians for multimodal data
  clustering}.
\newblock \bibinfo{journal}{\emph{IEEE transactions on pattern analysis and
  machine intelligence}} \bibinfo{volume}{43}, \bibinfo{number}{3}
  (\bibinfo{year}{2019}), \bibinfo{pages}{798--813}.
\newblock


\bibitem[Kumar et~al\mbox{.}(2011)]%
        {kumar2011co}
\bibfield{author}{\bibinfo{person}{Abhishek Kumar}, \bibinfo{person}{Piyush
  Rai}, {and} \bibinfo{person}{Hal Daume}.} \bibinfo{year}{2011}\natexlab{}.
\newblock \showarticletitle{Co-regularized multi-view spectral clustering}. In
  \bibinfo{booktitle}{\emph{Advances in neural information processing
  systems}}.
\newblock


\bibitem[Li et~al\mbox{.}(2022d)]%
        {LiangTNNLS}
\bibfield{author}{\bibinfo{person}{Liang Li}, \bibinfo{person}{Siwei Wang},
  \bibinfo{person}{Xinwang Liu}, \bibinfo{person}{En Zhu}, \bibinfo{person}{Li
  Shen}, \bibinfo{person}{Kenli Li}, {and} \bibinfo{person}{Keqin Li}.}
  \bibinfo{year}{2022}\natexlab{d}.
\newblock \showarticletitle{Local Sample-Weighted Multiple Kernel Clustering
  With Consensus Discriminative Graph}.
\newblock \bibinfo{journal}{\emph{IEEE Transactions on Neural Networks and
  Learning Systems}} (\bibinfo{year}{2022}), \bibinfo{pages}{1--14}.
\newblock
\urldef\tempurl%
\url{https://doi.org/10.1109/TNNLS.2022.3184970}
\showDOI{\tempurl}


\bibitem[Li et~al\mbox{.}(2023b)]%
        {LiangTKDE}
\bibfield{author}{\bibinfo{person}{Liang Li}, \bibinfo{person}{Junpu Zhang},
  \bibinfo{person}{Siwei Wang}, \bibinfo{person}{Xinwang Liu},
  \bibinfo{person}{Kenli Li}, {and} \bibinfo{person}{Keqin Li}.}
  \bibinfo{year}{2023}\natexlab{b}.
\newblock \showarticletitle{Multi-View Bipartite Graph Clustering With Coupled
  Noisy Feature Filter}.
\newblock \bibinfo{journal}{\emph{IEEE Transactions on Knowledge and Data
  Engineering}} (\bibinfo{year}{2023}), \bibinfo{pages}{1--13}.
\newblock
\urldef\tempurl%
\url{https://doi.org/10.1109/TKDE.2023.3268215}
\showDOI{\tempurl}


\bibitem[Li et~al\mbox{.}(2022c)]%
        {li2022parameter}
\bibfield{author}{\bibinfo{person}{Miaomiao Li}, \bibinfo{person}{Siwei Wang},
  \bibinfo{person}{Xinwang Liu}, {and} \bibinfo{person}{Suyuan Liu}.}
  \bibinfo{year}{2022}\natexlab{c}.
\newblock \showarticletitle{Parameter-Free and Scalable Incomplete Multiview
  Clustering With Prototype Graph}.
\newblock \bibinfo{journal}{\emph{IEEE Transactions on Neural Networks and
  Learning Systems}} (\bibinfo{year}{2022}).
\newblock


\bibitem[Li et~al\mbox{.}(2023a)]%
        {li2023dual}
\bibfield{author}{\bibinfo{person}{Qian Li}, \bibinfo{person}{Shu Guo},
  \bibinfo{person}{Cheng Ji}, \bibinfo{person}{Xutan Peng},
  \bibinfo{person}{Shiyao Cui}, {and} \bibinfo{person}{Jianxin Li}.}
  \bibinfo{year}{2023}\natexlab{a}.
\newblock \showarticletitle{Dual-Gated Fusion with Prefix-Tuning for
  Multi-Modal Relation Extraction}.
\newblock \bibinfo{journal}{\emph{arXiv preprint arXiv:2306.11020}}
  (\bibinfo{year}{2023}).
\newblock


\bibitem[Li et~al\mbox{.}(2019)]%
        {li2019flexible}
\bibfield{author}{\bibinfo{person}{Ruihuang Li}, \bibinfo{person}{Changqing
  Zhang}, \bibinfo{person}{Qinghua Hu}, \bibinfo{person}{Pengfei Zhu}, {and}
  \bibinfo{person}{Zheng Wang}.} \bibinfo{year}{2019}\natexlab{}.
\newblock \showarticletitle{Flexible Multi-View Representation Learning for
  Subspace Clustering.}. In \bibinfo{booktitle}{\emph{Proc. of IJCAI}}.
\newblock


\bibitem[Li et~al\mbox{.}(2014)]%
        {li2014partial}
\bibfield{author}{\bibinfo{person}{Shao-Yuan Li}, \bibinfo{person}{Yuan Jiang},
  {and} \bibinfo{person}{Zhi-Hua Zhou}.} \bibinfo{year}{2014}\natexlab{}.
\newblock \showarticletitle{Partial multi-view clustering}. In
  \bibinfo{booktitle}{\emph{Proc. of AAAI}}.
\newblock


\bibitem[Li et~al\mbox{.}(2022a)]%
        {li2022refining}
\bibfield{author}{\bibinfo{person}{Xiang-Long Li}, \bibinfo{person}{Man-Sheng
  Chen}, \bibinfo{person}{Chang-Dong Wang}, {and} \bibinfo{person}{Jian-Huang
  Lai}.} \bibinfo{year}{2022}\natexlab{a}.
\newblock \showarticletitle{Refining graph structure for incomplete multi-view
  clustering}.
\newblock \bibinfo{journal}{\emph{IEEE Transactions on Neural Networks and
  Learning Systems}} (\bibinfo{year}{2022}).
\newblock


\bibitem[Li et~al\mbox{.}(2018)]%
        {li2018survey}
\bibfield{author}{\bibinfo{person}{Yingming Li}, \bibinfo{person}{Ming Yang},
  {and} \bibinfo{person}{Zhongfei Zhang}.} \bibinfo{year}{2018}\natexlab{}.
\newblock \showarticletitle{A survey of multi-view representation learning}.
\newblock \bibinfo{journal}{\emph{IEEE transactions on knowledge and data
  engineering}} \bibinfo{volume}{31}, \bibinfo{number}{10}
  (\bibinfo{year}{2018}), \bibinfo{pages}{1863--1883}.
\newblock


\bibitem[Li et~al\mbox{.}(2022b)]%
        {li2022high}
\bibfield{author}{\bibinfo{person}{Zhenglai Li}, \bibinfo{person}{Chang Tang},
  \bibinfo{person}{Xiao Zheng}, \bibinfo{person}{Xinwang Liu},
  \bibinfo{person}{Wei Zhang}, {and} \bibinfo{person}{En Zhu}.}
  \bibinfo{year}{2022}\natexlab{b}.
\newblock \showarticletitle{High-order correlation preserved incomplete
  multi-view subspace clustering}.
\newblock \bibinfo{journal}{\emph{IEEE Transactions on Image Processing}}
  \bibinfo{volume}{31} (\bibinfo{year}{2022}), \bibinfo{pages}{2067--2080}.
\newblock


\bibitem[Liang et~al\mbox{.}(2022)]%
        {AKGR}
\bibfield{author}{\bibinfo{person}{Ke Liang}, \bibinfo{person}{Lingyuan Meng},
  \bibinfo{person}{Meng Liu}, \bibinfo{person}{Yue Liu},
  \bibinfo{person}{Wenxuan Tu}, \bibinfo{person}{Siwei Wang},
  \bibinfo{person}{Sihang Zhou}, \bibinfo{person}{X Liu}, {and}
  \bibinfo{person}{F Sun}.} \bibinfo{year}{2022}\natexlab{}.
\newblock \showarticletitle{A Survey of Knowledge Graph Reasoning on Graph
  Types: Static, Dynamic, and Multimodal}.
\newblock  (\bibinfo{year}{2022}).
\newblock


\bibitem[Liang et~al\mbox{.}(2021)]%
        {liang2021mka}
\bibfield{author}{\bibinfo{person}{Ke Liang}, \bibinfo{person}{Sifan Wu},
  \bibinfo{person}{Jiayi Gu}, {et~al\mbox{.}}} \bibinfo{year}{2021}\natexlab{}.
\newblock \showarticletitle{Mka: A scalable medical knowledge-assisted
  mechanism for generative models on medical conversation tasks}.
\newblock \bibinfo{journal}{\emph{Computational and mathematical methods in
  medicine}}  \bibinfo{volume}{2021} (\bibinfo{year}{2021}).
\newblock


\bibitem[Liang et~al\mbox{.}(2023)]%
        {liang2023structure}
\bibfield{author}{\bibinfo{person}{Ke Liang}, \bibinfo{person}{Sihang Zhou},
  \bibinfo{person}{Yue Liu}, \bibinfo{person}{Lingyuan Meng},
  \bibinfo{person}{Meng Liu}, {and} \bibinfo{person}{Xinwang Liu}.}
  \bibinfo{year}{2023}\natexlab{}.
\newblock \showarticletitle{Structure Guided Multi-modal Pre-trained
  Transformer for Knowledge Graph Reasoning}.
\newblock \bibinfo{journal}{\emph{arXiv preprint arXiv:2307.03591}}
  (\bibinfo{year}{2023}).
\newblock


\bibitem[Lin et~al\mbox{.}(2022a)]%
        {lin2022tensor}
\bibfield{author}{\bibinfo{person}{Jia-Qi Lin}, \bibinfo{person}{Man-Sheng
  Chen}, \bibinfo{person}{Chang-Dong Wang}, {and} \bibinfo{person}{Haizhang
  Zhang}.} \bibinfo{year}{2022}\natexlab{a}.
\newblock \showarticletitle{A Tensor Approach for Uncoupled Multiview
  Clustering}.
\newblock \bibinfo{journal}{\emph{IEEE Transactions on Cybernetics}}
  (\bibinfo{year}{2022}).
\newblock


\bibitem[Lin et~al\mbox{.}(2022c)]%
        {lin2022incomplete}
\bibfield{author}{\bibinfo{person}{Jia-Qi Lin}, \bibinfo{person}{Xiang-Long
  Li}, \bibinfo{person}{Man-Sheng Chen}, \bibinfo{person}{Chang-Dong Wang},
  {and} \bibinfo{person}{Haizhang Zhang}.} \bibinfo{year}{2022}\natexlab{c}.
\newblock \showarticletitle{Incomplete Data Meets Uncoupled Case: A Challenging
  Task of Multiview Clustering}.
\newblock \bibinfo{journal}{\emph{IEEE Transactions on Neural Networks and
  Learning Systems}} (\bibinfo{year}{2022}).
\newblock


\bibitem[Lin et~al\mbox{.}(2022b)]%
        {lin2022dual}
\bibfield{author}{\bibinfo{person}{Yijie Lin}, \bibinfo{person}{Yuanbiao Gou},
  \bibinfo{person}{Xiaotian Liu}, \bibinfo{person}{Jinfeng Bai},
  \bibinfo{person}{Jiancheng Lv}, {and} \bibinfo{person}{Xi Peng}.}
  \bibinfo{year}{2022}\natexlab{b}.
\newblock \showarticletitle{Dual contrastive prediction for incomplete
  multi-view representation learning}.
\newblock \bibinfo{journal}{\emph{IEEE Transactions on Pattern Analysis and
  Machine Intelligence}} (\bibinfo{year}{2022}).
\newblock


\bibitem[Lin et~al\mbox{.}(2021)]%
        {lin2021completer}
\bibfield{author}{\bibinfo{person}{Yijie Lin}, \bibinfo{person}{Yuanbiao Gou},
  \bibinfo{person}{Zitao Liu}, \bibinfo{person}{Boyun Li},
  \bibinfo{person}{Jiancheng Lv}, {and} \bibinfo{person}{Xi Peng}.}
  \bibinfo{year}{2021}\natexlab{}.
\newblock \showarticletitle{COMPLETER: Incomplete multi-view clustering via
  contrastive prediction}. In \bibinfo{booktitle}{\emph{Proc. of CVPR}}.
\newblock


\bibitem[Liu et~al\mbox{.}(2023b)]%
        {S2T_ML}
\bibfield{author}{\bibinfo{person}{Meng Liu}, \bibinfo{person}{Ke Liang},
  \bibinfo{person}{Bin Xiao}, \bibinfo{person}{Sihang Zhou},
  \bibinfo{person}{Wenxuan Tu}, \bibinfo{person}{Yue Liu},
  \bibinfo{person}{Xihong Yang}, {and} \bibinfo{person}{Xinwang Liu}.}
  \bibinfo{year}{2023}\natexlab{b}.
\newblock \showarticletitle{Self-Supervised Temporal Graph learning with
  Temporal and Structural Intensity Alignment}.
\newblock \bibinfo{journal}{\emph{arXiv preprint arXiv:2302.07491}}
  (\bibinfo{year}{2023}).
\newblock


\bibitem[Liu et~al\mbox{.}(2023c)]%
        {TGC_ML}
\bibfield{author}{\bibinfo{person}{Meng Liu}, \bibinfo{person}{Yue Liu},
  \bibinfo{person}{Ke Liang}, \bibinfo{person}{Siwei Wang},
  \bibinfo{person}{Sihang Zhou}, {and} \bibinfo{person}{Xinwang Liu}.}
  \bibinfo{year}{2023}\natexlab{c}.
\newblock \showarticletitle{Deep Temporal Graph Clustering}.
\newblock \bibinfo{journal}{\emph{arXiv preprint arXiv:2305.10738}}
  (\bibinfo{year}{2023}).
\newblock


\bibitem[Liu et~al\mbox{.}(2022a)]%
        {liu2022fast}
\bibfield{author}{\bibinfo{person}{Suyuan Liu}, \bibinfo{person}{Xinwang Liu},
  \bibinfo{person}{Siwei Wang}, \bibinfo{person}{Xin Niu}, {and}
  \bibinfo{person}{En Zhu}.} \bibinfo{year}{2022}\natexlab{a}.
\newblock \showarticletitle{Fast Incomplete Multi-View Clustering With
  View-Independent Anchors}.
\newblock \bibinfo{journal}{\emph{IEEE Transactions on Neural Networks and
  Learning Systems}} (\bibinfo{year}{2022}).
\newblock


\bibitem[Liu et~al\mbox{.}(2010)]%
        {liu2010large}
\bibfield{author}{\bibinfo{person}{Wei Liu}, \bibinfo{person}{Junfeng He},
  {and} \bibinfo{person}{Shih-Fu Chang}.} \bibinfo{year}{2010}\natexlab{}.
\newblock \showarticletitle{Large graph construction for scalable
  semi-supervised learning}. In \bibinfo{booktitle}{\emph{ICML}}.
\newblock


\bibitem[Liu et~al\mbox{.}(2020)]%
        {liu2020efficient}
\bibfield{author}{\bibinfo{person}{X Liu}, \bibinfo{person}{M Li},
  \bibinfo{person}{C Tang}, \bibinfo{person}{J Xia}, \bibinfo{person}{J Xiong},
  \bibinfo{person}{L Liu}, \bibinfo{person}{M Kloft}, {and} \bibinfo{person}{E
  Zhu}.} \bibinfo{year}{2020}\natexlab{}.
\newblock \showarticletitle{Efficient and Effective Regularized Incomplete
  Multi-view Clustering.}
\newblock \bibinfo{journal}{\emph{IEEE transactions on pattern analysis and
  machine intelligence}} (\bibinfo{year}{2020}).
\newblock


\bibitem[Liu et~al\mbox{.}(2017)]%
        {liu2017multiple}
\bibfield{author}{\bibinfo{person}{Xinwang Liu}, \bibinfo{person}{Miaomiao Li},
  \bibinfo{person}{Lei Wang}, \bibinfo{person}{Yong Dou},
  \bibinfo{person}{Jianping Yin}, {and} \bibinfo{person}{En Zhu}.}
  \bibinfo{year}{2017}\natexlab{}.
\newblock \showarticletitle{Multiple Kernel k-Means with Incomplete Kernels}.
  In \bibinfo{booktitle}{\emph{Thirty-First AAAI Conference on Artificial
  Intelligence}}.
\newblock


\bibitem[{Liu} et~al\mbox{.}(2020)]%
        {liu2020multiple}
\bibfield{author}{\bibinfo{person}{Xinwang {Liu}}, \bibinfo{person}{Xinzhong
  {Zhu}}, \bibinfo{person}{Miaomiao {Li}}, \bibinfo{person}{Lei {Wang}},
  \bibinfo{person}{En {Zhu}}, \bibinfo{person}{Tongliang {Liu}},
  \bibinfo{person}{Marius {Kloft}}, \bibinfo{person}{Dinggang {Shen}},
  \bibinfo{person}{Jianping {Yin}}, {and} \bibinfo{person}{Wen {Gao}}.}
  \bibinfo{year}{2020}\natexlab{}.
\newblock \showarticletitle{Multiple Kernel $k$ k -Means with Incomplete
  Kernels}.
\newblock \bibinfo{journal}{\emph{IEEE Transactions on Pattern Analysis and
  Machine Intelligence}} (\bibinfo{year}{2020}).
\newblock


\bibitem[Liu et~al\mbox{.}(2023a)]%
        {liuyue_Dink_net}
\bibfield{author}{\bibinfo{person}{Yue Liu}, \bibinfo{person}{Ke Liang},
  \bibinfo{person}{Jun Xia}, \bibinfo{person}{Sihang Zhou},
  \bibinfo{person}{Xihong Yang}, \bibinfo{person}{}, \bibinfo{person}{Xinwang
  Liu}, {and} \bibinfo{person}{Z.~Stan Li}.} \bibinfo{year}{2023}\natexlab{a}.
\newblock \showarticletitle{Dink-Net: Neural Clustering on Large Graphs}. In
  \bibinfo{booktitle}{\emph{Proc. of ICML}}.
\newblock


\bibitem[Liu et~al\mbox{.}(2022b)]%
        {liuyue_DCRN}
\bibfield{author}{\bibinfo{person}{Yue Liu}, \bibinfo{person}{Wenxuan Tu},
  \bibinfo{person}{Sihang Zhou}, \bibinfo{person}{Xinwang Liu},
  \bibinfo{person}{Linxuan Song}, \bibinfo{person}{Xihong Yang}, {and}
  \bibinfo{person}{En Zhu}.} \bibinfo{year}{2022}\natexlab{b}.
\newblock \showarticletitle{Deep Graph Clustering via Dual Correlation
  Reduction}. In \bibinfo{booktitle}{\emph{Proceedings of the AAAI Conference
  on Artificial Intelligence}}, Vol.~\bibinfo{volume}{36}.
  \bibinfo{pages}{7603--7611}.
\newblock


\bibitem[Liu et~al\mbox{.}(2022c)]%
        {liuyue_survey}
\bibfield{author}{\bibinfo{person}{Yue Liu}, \bibinfo{person}{Jun Xia},
  \bibinfo{person}{Sihang Zhou}, \bibinfo{person}{Siwei Wang},
  \bibinfo{person}{Xifeng Guo}, \bibinfo{person}{Xihong Yang},
  \bibinfo{person}{Ke Liang}, \bibinfo{person}{Wenxuan Tu},
  \bibinfo{person}{Z.~Stan Li}, {and} \bibinfo{person}{Xinwang Liu}.}
  \bibinfo{year}{2022}\natexlab{c}.
\newblock \showarticletitle{A Survey of Deep Graph Clustering: Taxonomy,
  Challenge, and Application}.
\newblock \bibinfo{journal}{\emph{arXiv preprint arXiv:2211.12875}}
  (\bibinfo{year}{2022}).
\newblock


\bibitem[Liu et~al\mbox{.}(2023d)]%
        {liuyue_SCGC}
\bibfield{author}{\bibinfo{person}{Yue Liu}, \bibinfo{person}{Xihong Yang},
  \bibinfo{person}{Sihang Zhou}, {and} \bibinfo{person}{Xinwang Liu}.}
  \bibinfo{year}{2023}\natexlab{d}.
\newblock \showarticletitle{Simple contrastive graph clustering}.
\newblock \bibinfo{journal}{\emph{IEEE Transactions on Neural Networks and
  Learning Systems}} (\bibinfo{year}{2023}).
\newblock


\bibitem[Liu et~al\mbox{.}(2023e)]%
        {liuyue_HSAN}
\bibfield{author}{\bibinfo{person}{Yue Liu}, \bibinfo{person}{Xihong Yang},
  \bibinfo{person}{Sihang Zhou}, \bibinfo{person}{Xinwang Liu},
  \bibinfo{person}{Zhen Wang}, \bibinfo{person}{Ke Liang},
  \bibinfo{person}{Wenxuan Tu}, \bibinfo{person}{Liang Li},
  \bibinfo{person}{Jingcan Duan}, {and} \bibinfo{person}{Cancan Chen}.}
  \bibinfo{year}{2023}\natexlab{e}.
\newblock \showarticletitle{Hard Sample Aware Network for Contrastive Deep
  Graph Clustering}. In \bibinfo{booktitle}{\emph{Proc. of AAAI}}.
\newblock


\bibitem[Mo et~al\mbox{.}(2022)]%
        {Mo_AAAI_2022}
\bibfield{author}{\bibinfo{person}{Yujie Mo}, \bibinfo{person}{Liang Peng},
  \bibinfo{person}{Jie Xu}, \bibinfo{person}{Xiaoshuang Shi}, {and}
  \bibinfo{person}{Xiaofeng Zhu}.} \bibinfo{year}{2022}\natexlab{}.
\newblock \showarticletitle{Simple Unsupervised Graph Representation Learning}.
  In \bibinfo{booktitle}{\emph{Proceedings of the AAAI Conference on Artificial
  Intelligence (AAAI)}}. \bibinfo{pages}{7797--7805}.
\newblock


\bibitem[Ng et~al\mbox{.}(2002)]%
        {ng2002spectral}
\bibfield{author}{\bibinfo{person}{Andrew~Y Ng}, \bibinfo{person}{Michael~I
  Jordan}, {and} \bibinfo{person}{Yair Weiss}.}
  \bibinfo{year}{2002}\natexlab{}.
\newblock \showarticletitle{On spectral clustering: Analysis and an algorithm}.
  In \bibinfo{booktitle}{\emph{Advances in neural information processing
  systems}}. \bibinfo{pages}{849--856}.
\newblock


\bibitem[Nie et~al\mbox{.}(2017)]%
        {nie2017auto}
\bibfield{author}{\bibinfo{person}{Feiping Nie}, \bibinfo{person}{Guohao Cai},
  \bibinfo{person}{Jing Li}, {and} \bibinfo{person}{Xuelong Li}.}
  \bibinfo{year}{2017}\natexlab{}.
\newblock \showarticletitle{Auto-weighted multi-view learning for image
  clustering and semi-supervised classification}.
\newblock \bibinfo{journal}{\emph{IEEE Transactions on Image Processing}}
  (\bibinfo{year}{2017}).
\newblock


\bibitem[Nie et~al\mbox{.}(2016)]%
        {nie2016parameter}
\bibfield{author}{\bibinfo{person}{Feiping Nie}, \bibinfo{person}{Jing Li},
  \bibinfo{person}{Xuelong Li}, {et~al\mbox{.}}}
  \bibinfo{year}{2016}\natexlab{}.
\newblock \showarticletitle{Parameter-free auto-weighted multiple graph
  learning: a framework for multiview clustering and semi-supervised
  classification.}. In \bibinfo{booktitle}{\emph{IJCAI}}.
  \bibinfo{pages}{1881--1887}.
\newblock


\bibitem[Nie et~al\mbox{.}(2018)]%
        {nie2018multiview}
\bibfield{author}{\bibinfo{person}{Feiping Nie}, \bibinfo{person}{Lai Tian},
  {and} \bibinfo{person}{Xuelong Li}.} \bibinfo{year}{2018}\natexlab{}.
\newblock \showarticletitle{Multiview clustering via adaptively weighted
  procrustes}. In \bibinfo{booktitle}{\emph{Proceedings of the 24th ACM SIGKDD
  international conference on knowledge discovery \& data mining}}.
  \bibinfo{pages}{2022--2030}.
\newblock


\bibitem[Peng et~al\mbox{.}(2019)]%
        {peng2019comic}
\bibfield{author}{\bibinfo{person}{Xi Peng}, \bibinfo{person}{Zhenyu Huang},
  \bibinfo{person}{Jiancheng Lv}, \bibinfo{person}{Hongyuan Zhu}, {and}
  \bibinfo{person}{Joey~Tianyi Zhou}.} \bibinfo{year}{2019}\natexlab{}.
\newblock \showarticletitle{COMIC: Multi-view clustering without parameter
  selection}. In \bibinfo{booktitle}{\emph{Proc. of ICML}}.
\newblock


\bibitem[Qiang et~al\mbox{.}(2021)]%
        {qiang2021fast}
\bibfield{author}{\bibinfo{person}{Qianyao Qiang}, \bibinfo{person}{Bin Zhang},
  \bibinfo{person}{Fei Wang}, {and} \bibinfo{person}{Feiping Nie}.}
  \bibinfo{year}{2021}\natexlab{}.
\newblock \showarticletitle{Fast multi-view discrete clustering with anchor
  graphs}. In \bibinfo{booktitle}{\emph{Proceedings of the AAAI Conference on
  Artificial Intelligence}}, Vol.~\bibinfo{volume}{35}.
  \bibinfo{pages}{9360--9367}.
\newblock


\bibitem[Ren et~al\mbox{.}(2019)]%
        {ren2019semi}
\bibfield{author}{\bibinfo{person}{Yazhou Ren}, \bibinfo{person}{Kangrong Hu},
  \bibinfo{person}{Xinyi Dai}, \bibinfo{person}{Lili Pan},
  \bibinfo{person}{Steven~CH Hoi}, {and} \bibinfo{person}{Zenglin Xu}.}
  \bibinfo{year}{2019}\natexlab{}.
\newblock \showarticletitle{Semi-supervised deep embedded clustering}.
\newblock \bibinfo{journal}{\emph{Neurocomputing}}  \bibinfo{volume}{325}
  (\bibinfo{year}{2019}), \bibinfo{pages}{121--130}.
\newblock


\bibitem[Ren et~al\mbox{.}(2020)]%
        {ren2020consensus}
\bibfield{author}{\bibinfo{person}{Zhenwen Ren}, \bibinfo{person}{Simon~X
  Yang}, \bibinfo{person}{Quansen Sun}, {and} \bibinfo{person}{Tao Wang}.}
  \bibinfo{year}{2020}\natexlab{}.
\newblock \showarticletitle{Consensus affinity graph learning for multiple
  kernel clustering}.
\newblock \bibinfo{journal}{\emph{IEEE Transactions on Cybernetics}}
  \bibinfo{volume}{51}, \bibinfo{number}{6} (\bibinfo{year}{2020}),
  \bibinfo{pages}{3273--3284}.
\newblock


\bibitem[Shao et~al\mbox{.}(2015)]%
        {shao2015multiple}
\bibfield{author}{\bibinfo{person}{Weixiang Shao}, \bibinfo{person}{Lifang He},
  {and} \bibinfo{person}{S~Yu Philip}.} \bibinfo{year}{2015}\natexlab{}.
\newblock \showarticletitle{Multiple incomplete views clustering via weighted
  nonnegative matrix factorization with $\ell_{2,1}$ regularization}. In
  \bibinfo{booktitle}{\emph{Joint European conference on machine learning and
  knowledge discovery in databases}}.
\newblock


\bibitem[Wan et~al\mbox{.}(2022)]%
        {10.1145/3503161.3547864}
\bibfield{author}{\bibinfo{person}{Xinhang Wan}, \bibinfo{person}{Jiyuan Liu},
  \bibinfo{person}{Weixuan Liang}, \bibinfo{person}{Xinwang Liu},
  \bibinfo{person}{Yi Wen}, {and} \bibinfo{person}{En Zhu}.}
  \bibinfo{year}{2022}\natexlab{}.
\newblock \showarticletitle{Continual Multi-View Clustering}. In
  \bibinfo{booktitle}{\emph{Proceedings of the 30th ACM International
  Conference on Multimedia}} (Lisboa, Portugal) \emph{(\bibinfo{series}{MM
  '22})}. \bibinfo{publisher}{Association for Computing Machinery},
  \bibinfo{address}{New York, NY, USA}, \bibinfo{pages}{3676–3684}.
\newblock
\showISBNx{9781450392037}
\urldef\tempurl%
\url{https://doi.org/10.1145/3503161.3547864}
\showDOI{\tempurl}


\bibitem[Wan et~al\mbox{.}(2023b)]%
        {wan2023onestep}
\bibfield{author}{\bibinfo{person}{Xinhang Wan}, \bibinfo{person}{Jiyuan Liu},
  \bibinfo{person}{Xinwang Liu}, \bibinfo{person}{Siwei Wang},
  \bibinfo{person}{Yi Wen}, \bibinfo{person}{Tianjiao Wan}, \bibinfo{person}{Li
  Shen}, {and} \bibinfo{person}{En Zhu}.} \bibinfo{year}{2023}\natexlab{b}.
\newblock \bibinfo{title}{One-step Multi-view Clustering with Diverse
  Representation}.
\newblock
\newblock
\showeprint[arxiv]{2306.05437}~[cs.LG]


\bibitem[Wan et~al\mbox{.}(2023a)]%
        {wan2023autoweighted}
\bibfield{author}{\bibinfo{person}{Xinhang Wan}, \bibinfo{person}{Xinwang Liu},
  \bibinfo{person}{Jiyuan Liu}, \bibinfo{person}{Siwei Wang},
  \bibinfo{person}{Yi Wen}, \bibinfo{person}{Weixuan Liang},
  \bibinfo{person}{En Zhu}, \bibinfo{person}{Zhe Liu}, {and}
  \bibinfo{person}{Lu Zhou}.} \bibinfo{year}{2023}\natexlab{a}.
\newblock \bibinfo{title}{Auto-weighted Multi-view Clustering for Large-scale
  Data}.
\newblock
\newblock
\showeprint[arxiv]{2303.01983}~[cs.LG]


\bibitem[Wan et~al\mbox{.}(2023c)]%
        {wan2023fast}
\bibfield{author}{\bibinfo{person}{Xinhang Wan}, \bibinfo{person}{Bin Xiao},
  \bibinfo{person}{Xinwang Liu}, \bibinfo{person}{Jiyuan Liu},
  \bibinfo{person}{Weixuan Liang}, {and} \bibinfo{person}{En Zhu}.}
  \bibinfo{year}{2023}\natexlab{c}.
\newblock \bibinfo{title}{Fast Continual Multi-View Clustering with Incomplete
  Views}.
\newblock
\newblock
\showeprint[arxiv]{2306.02389}~[cs.LG]


\bibitem[Wang et~al\mbox{.}(2020a)]%
        {wang2020smoothness}
\bibfield{author}{\bibinfo{person}{Chang-Dong Wang}, \bibinfo{person}{Man-Sheng
  Chen}, \bibinfo{person}{Ling Huang}, \bibinfo{person}{Jian-Huang Lai}, {and}
  \bibinfo{person}{S~Yu Philip}.} \bibinfo{year}{2020}\natexlab{a}.
\newblock \showarticletitle{Smoothness regularized multiview subspace
  clustering with kernel learning}.
\newblock \bibinfo{journal}{\emph{IEEE Transactions on Neural Networks and
  Learning Systems}} (\bibinfo{year}{2020}).
\newblock


\bibitem[Wang et~al\mbox{.}(2019b)]%
        {wang2019gmc}
\bibfield{author}{\bibinfo{person}{Hao Wang}, \bibinfo{person}{Yan Yang}, {and}
  \bibinfo{person}{Bing Liu}.} \bibinfo{year}{2019}\natexlab{b}.
\newblock \showarticletitle{GMC: Graph-based multi-view clustering}.
\newblock \bibinfo{journal}{\emph{IEEE Transactions on Knowledge and Data
  Engineering}} (\bibinfo{year}{2019}).
\newblock


\bibitem[Wang et~al\mbox{.}(2019c)]%
        {wang2019spectral}
\bibfield{author}{\bibinfo{person}{Hao Wang}, \bibinfo{person}{Linlin Zong},
  \bibinfo{person}{Bing Liu}, \bibinfo{person}{Yan Yang}, {and}
  \bibinfo{person}{Wei Zhou}.} \bibinfo{year}{2019}\natexlab{c}.
\newblock \showarticletitle{Spectral perturbation meets incomplete multi-view
  data}.
\newblock \bibinfo{journal}{\emph{arXiv preprint arXiv:1906.00098}}
  (\bibinfo{year}{2019}).
\newblock


\bibitem[Wang et~al\mbox{.}(2018)]%
        {wang2018partial}
\bibfield{author}{\bibinfo{person}{Qianqian Wang}, \bibinfo{person}{Zhengming
  Ding}, \bibinfo{person}{Zhiqiang Tao}, \bibinfo{person}{Quanxue Gao}, {and}
  \bibinfo{person}{Yun Fu}.} \bibinfo{year}{2018}\natexlab{}.
\newblock \showarticletitle{Partial multi-view clustering via consistent GAN}.
  In \bibinfo{booktitle}{\emph{Proc. of ICDM}}.
\newblock


\bibitem[Wang et~al\mbox{.}(2020b)]%
        {wang2020icmsc}
\bibfield{author}{\bibinfo{person}{Qianqian Wang}, \bibinfo{person}{Huanhuan
  Lian}, \bibinfo{person}{Gan Sun}, \bibinfo{person}{Quanxue Gao}, {and}
  \bibinfo{person}{Licheng Jiao}.} \bibinfo{year}{2020}\natexlab{b}.
\newblock \showarticletitle{ICMSC: Incomplete cross-modal subspace clustering}.
\newblock \bibinfo{journal}{\emph{IEEE Transactions on Image Processing}}
  \bibinfo{volume}{30} (\bibinfo{year}{2020}), \bibinfo{pages}{305--317}.
\newblock


\bibitem[Wang et~al\mbox{.}(2022)]%
        {wang2022align}
\bibfield{author}{\bibinfo{person}{Siwei Wang}, \bibinfo{person}{Xinwang Liu},
  \bibinfo{person}{Suyuan Liu}, \bibinfo{person}{Jiaqi Jin},
  \bibinfo{person}{Wenxuan Tu}, \bibinfo{person}{Xinzhong Zhu}, {and}
  \bibinfo{person}{En Zhu}.} \bibinfo{year}{2022}\natexlab{}.
\newblock \showarticletitle{Align then Fusion: Generalized Large-scale
  Multi-view Clustering with Anchor Matching Correspondences}.
\newblock \bibinfo{journal}{\emph{arXiv preprint arXiv:2205.15075}}
  (\bibinfo{year}{2022}).
\newblock


\bibitem[Wang et~al\mbox{.}(2019a)]%
        {wang2019multi}
\bibfield{author}{\bibinfo{person}{Siwei Wang}, \bibinfo{person}{Xinwang Liu},
  \bibinfo{person}{En Zhu}, \bibinfo{person}{Chang Tang},
  \bibinfo{person}{Jiyuan Liu}, \bibinfo{person}{Jingtao Hu},
  \bibinfo{person}{Jingyuan Xia}, {and} \bibinfo{person}{Jianping Yin}.}
  \bibinfo{year}{2019}\natexlab{a}.
\newblock \showarticletitle{Multi-view Clustering via Late Fusion Alignment
  Maximization.}. In \bibinfo{booktitle}{\emph{IJCAI}}.
  \bibinfo{pages}{3778--3784}.
\newblock


\bibitem[Wang(2021)]%
        {wang2021survey}
\bibfield{author}{\bibinfo{person}{Yang Wang}.}
  \bibinfo{year}{2021}\natexlab{}.
\newblock \showarticletitle{Survey on deep multi-modal data analytics:
  Collaboration, rivalry, and fusion}.
\newblock \bibinfo{journal}{\emph{ACM Transactions on Multimedia Computing,
  Communications, and Applications (TOMM)}} \bibinfo{volume}{17},
  \bibinfo{number}{1s} (\bibinfo{year}{2021}), \bibinfo{pages}{1--25}.
\newblock


\bibitem[Wen et~al\mbox{.}(2021)]%
        {wen2021structural}
\bibfield{author}{\bibinfo{person}{Jie Wen}, \bibinfo{person}{Zhihao Wu},
  \bibinfo{person}{Zheng Zhang}, \bibinfo{person}{Lunke Fei},
  \bibinfo{person}{Bob Zhang}, {and} \bibinfo{person}{Yong Xu}.}
  \bibinfo{year}{2021}\natexlab{}.
\newblock \showarticletitle{Structural deep incomplete multi-view clustering
  network}. In \bibinfo{booktitle}{\emph{Proceedings of the 30th ACM
  International Conference on Information \& Knowledge Management}}.
  \bibinfo{pages}{3538--3542}.
\newblock


\bibitem[Wen et~al\mbox{.}(2019)]%
        {wen2019unified}
\bibfield{author}{\bibinfo{person}{Jie Wen}, \bibinfo{person}{Zheng Zhang},
  \bibinfo{person}{Yong Xu}, \bibinfo{person}{Bob Zhang},
  \bibinfo{person}{Lunke Fei}, {and} \bibinfo{person}{Hong Liu}.}
  \bibinfo{year}{2019}\natexlab{}.
\newblock \showarticletitle{Unified embedding alignment with missing views
  inferring for incomplete multi-view clustering}. In
  \bibinfo{booktitle}{\emph{Proc. of AAAI}}.
\newblock


\bibitem[Wen et~al\mbox{.}(2020a)]%
        {wen2020cdimc}
\bibfield{author}{\bibinfo{person}{Jie Wen}, \bibinfo{person}{Zheng Zhang},
  \bibinfo{person}{Yong Xu}, \bibinfo{person}{Bob Zhang},
  \bibinfo{person}{Lunke Fei}, {and} \bibinfo{person}{Guo-Sen Xie}.}
  \bibinfo{year}{2020}\natexlab{a}.
\newblock \showarticletitle{CDIMC-net: Cognitive Deep Incomplete Multi-view
  Clustering Network.}. In \bibinfo{booktitle}{\emph{Proc. of IJCAI}}.
\newblock


\bibitem[Wen et~al\mbox{.}(2020b)]%
        {wen2020generalized}
\bibfield{author}{\bibinfo{person}{Jie Wen}, \bibinfo{person}{Zheng Zhang},
  \bibinfo{person}{Zhao Zhang}, \bibinfo{person}{Lunke Fei}, {and}
  \bibinfo{person}{Meng Wang}.} \bibinfo{year}{2020}\natexlab{b}.
\newblock \showarticletitle{Generalized incomplete multiview clustering with
  flexible locality structure diffusion}.
\newblock \bibinfo{journal}{\emph{IEEE transactions on cybernetics}}
  (\bibinfo{year}{2020}).
\newblock


\bibitem[Xu et~al\mbox{.}(2023)]%
        {xu2023adaptive}
\bibfield{author}{\bibinfo{person}{Jie Xu}, \bibinfo{person}{Chao Li},
  \bibinfo{person}{Liang Peng}, \bibinfo{person}{Yazhou Ren},
  \bibinfo{person}{Xiaoshuang Shi}, \bibinfo{person}{Heng~Tao Shen}, {and}
  \bibinfo{person}{Xiaofeng Zhu}.} \bibinfo{year}{2023}\natexlab{}.
\newblock \showarticletitle{Adaptive Feature Projection With Distribution
  Alignment for Deep Incomplete Multi-View Clustering}.
\newblock \bibinfo{journal}{\emph{IEEE Transactions on Image Processing}}
  \bibinfo{volume}{32} (\bibinfo{year}{2023}), \bibinfo{pages}{1354--1366}.
\newblock


\bibitem[Xu et~al\mbox{.}(2022)]%
        {xu2022multi}
\bibfield{author}{\bibinfo{person}{Jie Xu}, \bibinfo{person}{Huayi Tang},
  \bibinfo{person}{Yazhou Ren}, \bibinfo{person}{Liang Peng},
  \bibinfo{person}{Xiaofeng Zhu}, {and} \bibinfo{person}{Lifang He}.}
  \bibinfo{year}{2022}\natexlab{}.
\newblock \showarticletitle{Multi-level feature learning for contrastive
  multi-view clustering}. In \bibinfo{booktitle}{\emph{Proceedings of the
  IEEE/CVF Conference on Computer Vision and Pattern Recognition}}.
  \bibinfo{pages}{16051--16060}.
\newblock


\bibitem[Yang et~al\mbox{.}(2022a)]%
        {yang2022robust}
\bibfield{author}{\bibinfo{person}{Mouxing Yang}, \bibinfo{person}{Yunfan Li},
  \bibinfo{person}{Peng Hu}, \bibinfo{person}{Jinfeng Bai},
  \bibinfo{person}{Jiancheng Lv}, {and} \bibinfo{person}{Xi Peng}.}
  \bibinfo{year}{2022}\natexlab{a}.
\newblock \showarticletitle{Robust multi-view clustering with incomplete
  information}.
\newblock \bibinfo{journal}{\emph{IEEE Transactions on Pattern Analysis and
  Machine Intelligence}} \bibinfo{volume}{45}, \bibinfo{number}{1}
  (\bibinfo{year}{2022}), \bibinfo{pages}{1055--1069}.
\newblock


\bibitem[Yang et~al\mbox{.}(2021)]%
        {yang2021partially}
\bibfield{author}{\bibinfo{person}{Mouxing Yang}, \bibinfo{person}{Yunfan Li},
  \bibinfo{person}{Zhenyu Huang}, \bibinfo{person}{Zitao Liu},
  \bibinfo{person}{Peng Hu}, {and} \bibinfo{person}{Xi Peng}.}
  \bibinfo{year}{2021}\natexlab{}.
\newblock \showarticletitle{Partially view-aligned representation learning with
  noise-robust contrastive loss}. In \bibinfo{booktitle}{\emph{Proceedings of
  the IEEE/CVF conference on computer vision and pattern recognition}}.
  \bibinfo{pages}{1134--1143}.
\newblock


\bibitem[Yang et~al\mbox{.}(2022b)]%
        {MGCN}
\bibfield{author}{\bibinfo{person}{Xihong Yang}, \bibinfo{person}{Yue Liu},
  \bibinfo{person}{Sihang Zhou}, \bibinfo{person}{Xinwang Liu}, {and}
  \bibinfo{person}{En Zhu}.} \bibinfo{year}{2022}\natexlab{b}.
\newblock \showarticletitle{Mixed Graph Contrastive Network for Semi-Supervised
  Node Classification}.
\newblock \bibinfo{journal}{\emph{arXiv preprint arXiv:2206.02796}}
  (\bibinfo{year}{2022}).
\newblock


\bibitem[Yang et~al\mbox{.}(2022c)]%
        {GCC-LDA}
\bibfield{author}{\bibinfo{person}{Xihong Yang}, \bibinfo{person}{Yue Liu},
  \bibinfo{person}{Sihang Zhou}, \bibinfo{person}{Siwei Wang},
  \bibinfo{person}{Xinwang Liu}, {and} \bibinfo{person}{En Zhu}.}
  \bibinfo{year}{2022}\natexlab{c}.
\newblock \showarticletitle{Contrastive Deep Graph Clustering with Learnable
  Augmentation}.
\newblock \bibinfo{journal}{\emph{arXiv preprint arXiv:2212.03559}}
  (\bibinfo{year}{2022}).
\newblock


\bibitem[Yang et~al\mbox{.}(2023)]%
        {CCGC}
\bibfield{author}{\bibinfo{person}{Xihong Yang}, \bibinfo{person}{Yue Liu},
  \bibinfo{person}{Sihang Zhou}, \bibinfo{person}{Siwei Wang},
  \bibinfo{person}{Wenxuan Tu}, \bibinfo{person}{Qun Zheng},
  \bibinfo{person}{Xinwang Liu}, \bibinfo{person}{Liming Fang}, {and}
  \bibinfo{person}{En Zhu}.} \bibinfo{year}{2023}\natexlab{}.
\newblock \showarticletitle{Cluster-guided Contrastive Graph Clustering
  Network}.
\newblock \bibinfo{journal}{\emph{arXiv preprint arXiv:2301.01098}}
  (\bibinfo{year}{2023}).
\newblock


\bibitem[Zhan et~al\mbox{.}(2018)]%
        {zhan2018multiview}
\bibfield{author}{\bibinfo{person}{Kun Zhan}, \bibinfo{person}{Feiping Nie},
  \bibinfo{person}{Jing Wang}, {and} \bibinfo{person}{Yi Yang}.}
  \bibinfo{year}{2018}\natexlab{}.
\newblock \showarticletitle{Multiview consensus graph clustering}.
\newblock \bibinfo{journal}{\emph{IEEE Transactions on Image Processing}}
  (\bibinfo{year}{2018}).
\newblock


\bibitem[Zhan et~al\mbox{.}(2017)]%
        {zhan2017graph}
\bibfield{author}{\bibinfo{person}{Kun Zhan}, \bibinfo{person}{Changqing
  Zhang}, \bibinfo{person}{Junpeng Guan}, {and} \bibinfo{person}{Junsheng
  Wang}.} \bibinfo{year}{2017}\natexlab{}.
\newblock \showarticletitle{Graph learning for multiview clustering}.
\newblock \bibinfo{journal}{\emph{IEEE transactions on cybernetics}}
  \bibinfo{volume}{48}, \bibinfo{number}{10} (\bibinfo{year}{2017}),
  \bibinfo{pages}{2887--2895}.
\newblock


\bibitem[Zhang et~al\mbox{.}(2017)]%
        {zhang2017latent}
\bibfield{author}{\bibinfo{person}{Changqing Zhang}, \bibinfo{person}{Qinghua
  Hu}, \bibinfo{person}{Huazhu Fu}, \bibinfo{person}{Pengfei Zhu}, {and}
  \bibinfo{person}{Xiaochun Cao}.} \bibinfo{year}{2017}\natexlab{}.
\newblock \showarticletitle{Latent multi-view subspace clustering}. In
  \bibinfo{booktitle}{\emph{Proc. of CVPR}}.
\newblock


\bibitem[Zhang et~al\mbox{.}(2022)]%
        {ZJPACMMM}
\bibfield{author}{\bibinfo{person}{Junpu Zhang}, \bibinfo{person}{Liang Li},
  \bibinfo{person}{Siwei Wang}, \bibinfo{person}{Jiyuan Liu},
  \bibinfo{person}{Yue Liu}, \bibinfo{person}{Xinwang Liu}, {and}
  \bibinfo{person}{En Zhu}.} \bibinfo{year}{2022}\natexlab{}.
\newblock \showarticletitle{Multiple Kernel Clustering with Dual Noise
  Minimization}. In \bibinfo{booktitle}{\emph{Proceedings of the 30th ACM
  International Conference on Multimedia}} (Lisboa, Portugal)
  \emph{(\bibinfo{series}{MM '22})}. \bibinfo{publisher}{Association for
  Computing Machinery}, \bibinfo{address}{New York, NY, USA},
  \bibinfo{pages}{3440–3450}.
\newblock
\showISBNx{9781450392037}
\urldef\tempurl%
\url{https://doi.org/10.1145/3503161.3548334}
\showDOI{\tempurl}


\bibitem[Zhao et~al\mbox{.}(2017)]%
        {zhao2017multi}
\bibfield{author}{\bibinfo{person}{Handong Zhao}, \bibinfo{person}{Zhengming
  Ding}, {and} \bibinfo{person}{Yun Fu}.} \bibinfo{year}{2017}\natexlab{}.
\newblock \showarticletitle{Multi-view clustering via deep matrix
  factorization}. In \bibinfo{booktitle}{\emph{Proc. of AAAI}}.
\newblock


\bibitem[Zhao et~al\mbox{.}(2016)]%
        {zhao2016incomplete}
\bibfield{author}{\bibinfo{person}{Handong Zhao}, \bibinfo{person}{Hongfu Liu},
  {and} \bibinfo{person}{Yun Fu}.} \bibinfo{year}{2016}\natexlab{}.
\newblock \showarticletitle{Incomplete multi-modal visual data grouping.}. In
  \bibinfo{booktitle}{\emph{Proc. of IJCAI}}.
\newblock


\bibitem[Zhao et~al\mbox{.}(2022)]%
        {zhao2022mose}
\bibfield{author}{\bibinfo{person}{Yu Zhao}, \bibinfo{person}{Xiangrui Cai},
  \bibinfo{person}{Yike Wu}, \bibinfo{person}{Haiwei Zhang},
  \bibinfo{person}{Ying Zhang}, \bibinfo{person}{Guoqing Zhao}, {and}
  \bibinfo{person}{Ning Jiang}.} \bibinfo{year}{2022}\natexlab{}.
\newblock \showarticletitle{Mose: Modality split and ensemble for multimodal
  knowledge graph completion}.
\newblock \bibinfo{journal}{\emph{arXiv preprint arXiv:2210.08821}}
  (\bibinfo{year}{2022}).
\newblock


\bibitem[Zhu et~al\mbox{.}(2019)]%
        {zhu2019multi}
\bibfield{author}{\bibinfo{person}{Pengfei Zhu}, \bibinfo{person}{Binyuan Hui},
  \bibinfo{person}{Changqing Zhang}, \bibinfo{person}{Dawei Du},
  \bibinfo{person}{Longyin Wen}, {and} \bibinfo{person}{Qinghua Hu}.}
  \bibinfo{year}{2019}\natexlab{}.
\newblock \showarticletitle{Multi-view deep subspace clustering networks}.
\newblock \bibinfo{journal}{\emph{arXiv preprint arXiv:1908.01978}}
  (\bibinfo{year}{2019}).
\newblock


\end{thebibliography}


\end{document}